\documentclass{article}

 \usepackage[main, final]{neurips_2026}

\usepackage[utf8]{inputenc} %
\usepackage[T1]{fontenc}    %
\usepackage[pagebackref,breaklinks,colorlinks,citecolor=cvprblue]{hyperref}       %
\usepackage{url}            %
\usepackage{booktabs}       %
\usepackage{amsfonts}       %
\usepackage{nicefrac}       %
\usepackage{microtype}      %
\usepackage{xcolor}         %
\definecolor{cvprblue}{rgb}{0.21,0.49,0.74}
\usepackage{graphicx}
\usepackage{amsmath}
\usepackage{xspace}
\usepackage{bm}
\usepackage{nicematrix}
\usepackage{multirow}
\usepackage{caption}
\usepackage{float}
\newcommand{\ie}{\emph{i.e.}\xspace}
\newcommand{\eg}{\emph{e.g.}\xspace}

\title{OmniEgoCap: Device-Agnostic Sequence-Level Egocentric Motion Reconstruction}
\def\modelname{OmniEgoCap\xspace}

\author{
  Kyungwon Cho \qquad Jeonghyeon Na \qquad Hanbyul Joo \\
  [1.1em]
  Seoul National University \\
  [0.3em]
  {\tt\small \{cscandkswon,prom317,hbjoo\}@snu.ac.kr} \\
  [0.3em]
  {\tt\small \href{https://kyungwoncho.github.io/OmniEgoCap/}{\color{magenta}{https://kyungwoncho.github.io/OmniEgoCap/}}}
  \vspace{-5mm}
}

\begin{document}

\maketitle

\begin{figure}[H]\centering
\includegraphics[width=\linewidth, trim={0 0.5cm 0 0.0cm}, clip]{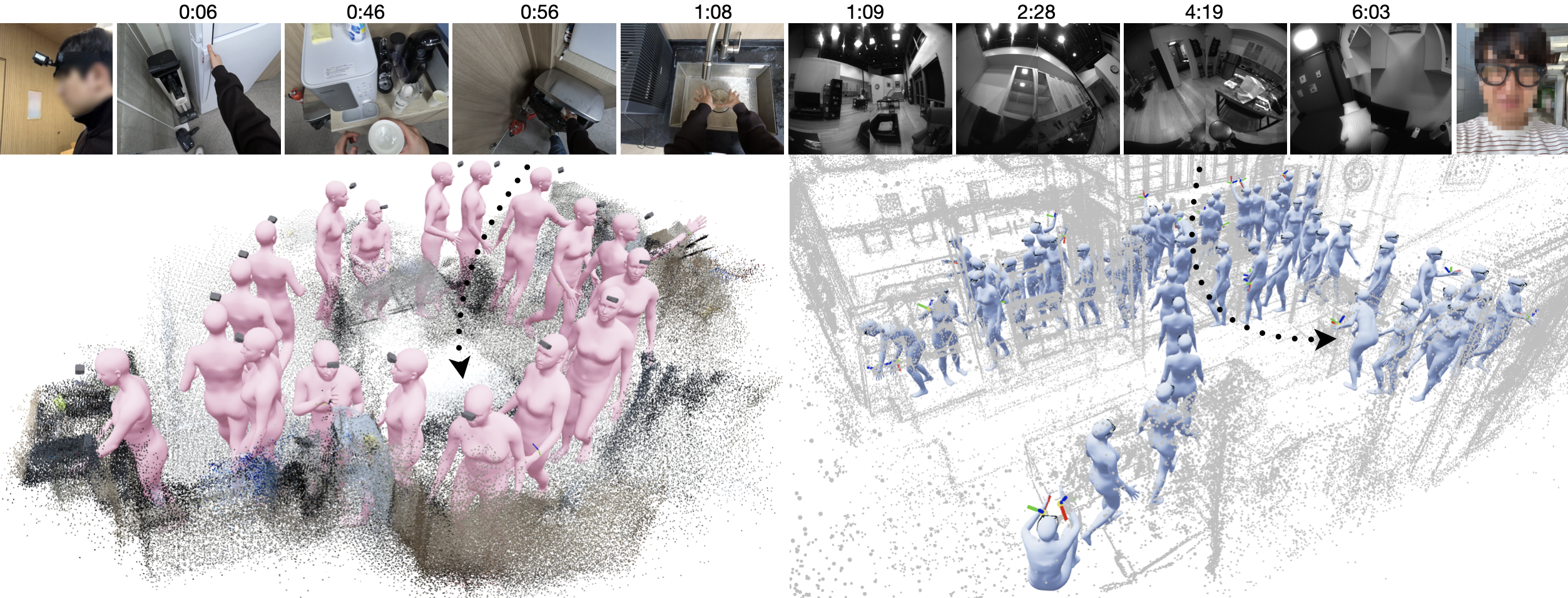}
\caption{
\textbf{\modelname} is a sequence-level diffusion framework that combines long-range body and camera-observation invariants with geometry-aware visibility augmentation to reconstruct consistent 3D full-body motion across diverse egocentric devices. (L) GoPro and (R) Aria.
}
\label{fig:teaser}
\end{figure}

\begin{abstract}
Commercial egocentric devices capture human behavior in everyday settings, yet full-body motion reconstruction must generalize across diverse cameras and mountings. This is challenging because head trajectories under-determine body motion, while hand observations are intermittent and camera-dependent: a hand may disappear simply because it has left the camera's field of view. Existing methods treat such absences as missing data, relying on fixed visibility regimes or post-hoc optimization, and fail to generalize across devices.
We present \modelname, a sequence-level diffusion framework for device-agnostic egocentric full-body motion reconstruction.
Our key insight is that egocentric videos contain sequence-level evidence about both the wearer and the camera-induced visibility structure, enabling \modelname to infer consistent body shape and coherent motion from sparse head and hand cues.
To prevent overfitting to a single camera setup, we further introduce geometry-aware visibility augmentation, which synthesizes visibility patterns from realistic variations in camera geometry.
We also present OmniEgoDB, the first mocap benchmark that pairs ground-truth motion with recordings from multiple consumer egocentric devices. Experiments on synthetic, real-device, and in-the-wild settings demonstrate superior motion reconstruction, robust cross-device generalization, and coherent in-the-wild results.
\end{abstract}

\section{Introduction}
Egocentric vision systems are rapidly diversifying, from action cameras~\cite{gopro} and VR headsets~\cite{metaquest,applevisionpro,samsungxr} to smart glasses~\cite{metarayban,metaaria,spectacles,vuzix}.
For machines to better assist humans, these systems need to understand 3D human motion and behavior. Yet reconstructing the wearer's 3D body from egocentric observations is challenging, since the first-person view rarely captures the full body.
Existing methods therefore rely on sparse cues, primarily head trajectories from SLAM or SfM~\cite{li2023egoego,yi2025egoallo} and visible hands from XR device tracking~\cite{metaaria,metaquest,applevisionpro,samsungxr} or off-the-shelf estimators~\cite{zhang2025hawor}.
As these devices become commonplace, full-body motion capture should be democratized across the devices people actually wear, ideally without device-specific retraining. This goal is complicated by the fact that hand visibility is itself camera-dependent: hands may disappear due to occlusion, tracking failure, or simply moving outside the camera's visible region. For device-agnostic reconstruction, hand observations must therefore be understood in relation to the camera geometry that produces them.

\looseness=-1
Existing methods do not fully exploit the camera geometry encoded in hand observations.
A missing hand is not merely absent evidence; across time, visibility and invisibility reveal what the camera can and cannot see. 
Optimization-based methods impose costly post-hoc constraints on visible hands~\cite{lee2024mocapevery,yi2025egoallo}, but do not model sustained invisibility as part of this observation process.
Recent end-to-end methods use head and hand signals directly~\cite{chi2024estimatingegobodyposedoubly,jiang2024egoposer,patel2025uniegomotion,guzov-jiang2025hmd2}, but are typically trained under a fixed camera setup.
They can therefore learn setup-specific visibility patterns and degrade when the setup changes.
These failures point to a common missing ingredient, sequence-level context.
An egocentric sequence contains long-range evidence about both \emph{who the wearer is} and \emph{how the camera observes them}.
Sparse transitions between visible and invisible hand states reveal the camera's observation boundary over time.
Body shape is also a sequence-level property.
It differs across wearers but is fixed within a sequence, while reliable evidence for height and proportions appears unevenly over time, especially in informative poses such as standing.
It is thus insufficient to rely on a fixed mean body shape~\cite{li2023egoego,lee2024mocapevery,guzov-jiang2025hmd2,jiang2022avatarposer,castillo2023bodiffusion} or to predict shape per frame~\cite{yi2025egoallo,patel2025uniegomotion}.
Thus, device-agnostic reconstruction requires inferring both the wearer's body shape and the camera's visibility context from the entire sequence.

Building on this view, we present \textbf{\modelname}, a sequence-level diffusion framework for device-agnostic egocentric full-body motion reconstruction.
It uses long-range cues to estimate body shape and a visibility context that captures camera-dependent observation structure, guiding motion reconstruction from sparse head and hand signals.
Our encoder separates sequence context from local motion evidence over long sequences, while a diffusion decoder reconstructs full-body motion conditioned on the inferred context and intermittent hand observations.
To make the model robust to diverse camera setups, we introduce \emph{geometry-aware visibility augmentation}.
Rather than randomly dropping hand observations, we synthesize diverse yet plausible visibility patterns induced by realistic variations in egocentric camera geometry and mounting, encouraging the model to treat both hand visibility and invisibility as geometric observation cues.
Evaluating this capability, however, is hindered by the fact that existing egocentric motion datasets are captured almost exclusively with a single device, most commonly Project Aria glasses~\cite{metaaria}, whose specialized hardware is not readily available to general consumers. We therefore introduce \textbf{OmniEgoDB}, the first public mocap benchmark that pairs ground-truth motion with recordings from multiple egocentric devices, including widely accessible consumer cameras and XR headsets.
Experiments on public benchmarks, OmniEgoDB, and in-the-wild videos demonstrate robust generalization across device categories, mountings, and visibility patterns.

In summary, our contributions are threefold.
(1) We propose a sequence-level diffusion framework that reconstructs long-range full-body motion from sparse egocentric cues while jointly inferring the wearer's body shape and a visibility context capturing camera-dependent observation structure.
(2) We introduce geometry-aware visibility augmentation, which models hand visibility as a camera-induced observation process, prevents overfitting to a fixed visibility regime, and enables robust cross-device generalization.
(3) We contribute a public mocap evaluation benchmark collected with multiple egocentric devices and mounting configurations, and demonstrate device-agnostic reconstruction without device-specific retraining.

\section{Related Work}

\paragraph{Sparse Observations for Egocentric Motion Capture}
Wearable IMU systems~\cite{xsens, rokoko, von2017sip, huang2018dip, yi2021transpose, yi2022pip, jiang2022tip, yi2024pnp, zhang2023dynamic, van2024diffusionposer} and VR/AR setups with 6-DoF head and hand tracking~\cite{jiang2022avatarposer, zheng2023realistic, dai2024hmd, jiang2024egoposer, aliakbarian2023hmdnemo, barquero2025sparse, dittadi2021fullbody, du2023agrol, castillo2023bodiffusion, tang2024unified, feng2024stratified} provide sparse yet informative body cues but rely on dedicated sensors, controllers, or global localization to reduce root drift~\cite{guzov2021hps, yi2023egolocate, lee2024mocapevery}.
Egocentric cameras further relax hardware requirements, but body visibility is severely limited.
Some methods improve visibility using specialized downward-facing or stereo fisheye cameras~\cite{rhodin2016egocap,wang2021estimating,jiang2021egocentric,tome2020selfpose,wang2023scene,akada2022unrealego,akada20243d,yang2024egoposeformer,wang2024egocentric,akada2025bring,lee2025rewind}, while more practical front-facing methods infer motion from head trajectories~\cite{li2023egoego}, visual features~\cite{patel2025uniegomotion, guzov-jiang2025hmd2}, or intermittent hand observations~\cite{chi2024estimatingegobodyposedoubly, yi2025egoallo}.
However, existing approaches usually assume a fixed sensing or visibility regime~\cite{chi2024estimatingegobodyposedoubly,jiang2024egoposer}, or use hand observations mainly through test-time optimization~\cite{yi2025egoallo,lee2024mocapevery}.
This limits their ability to model how hand visibility, including sustained disappearance, changes with camera geometry.
In contrast, we target \emph{device-agnostic} reconstruction from diverse head-mounted videos, where visible hands and sustained invisibility provide \emph{sequence-level evidence} of camera-induced visibility structure.

\paragraph{Generative Priors and Sequence-Level Modeling}
Diffusion models provide strong generative priors for high-fidelity motion synthesis~\cite{tevet2023human,jiang2023motiondiffusercontrollablemultiagentmotion,dabral2022mofusion,zhang2022motiondiffuse}, making them well suited to egocentric reconstruction where sparse head and hand cues leave multiple full-body motions plausible.
However, sparse local cues alone do not provide enough context to infer body shape or camera-dependent visibility structure.
Prior methods often assume a fixed mean shape~\cite{li2023egoego,lee2024mocapevery,guzov-jiang2025hmd2,jiang2022avatarposer,castillo2023bodiffusion} or predict shape locally~\cite{yi2025egoallo,patel2025uniegomotion}, leading to inconsistent body proportions and ambiguous motion interpretation.
As with body shape, camera-dependent visibility structure must be inferred from long-range evidence such as when and where hands appear and disappear.
Modeling such long-range evidence with transformer-based diffusion is challenging because standard attention scales quadratically with sequence length.
We adopt sliding-window attention~\cite{li2025genmo, barquero2024seamless} for efficient long-sequence denoising, while aggregating sequence-level context for the wearer's body shape and the camera's observation boundary.

\section{Method}
\begin{figure}[t]\centering
\includegraphics[width=\linewidth, trim={0 0 0 0},clip]{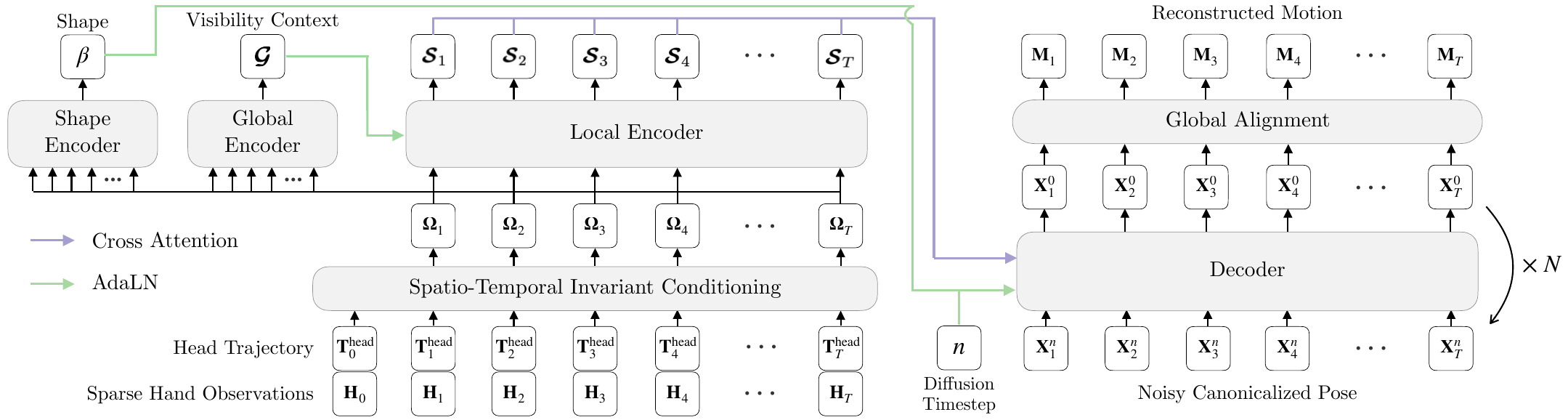}
\caption{\textbf{Overview of our model.} \modelname takes the head trajectory $\mathbf{T}^\text{head}_{0:T}$ and intermittent hand observations $\mathbf{H}_{0:T}$, which are first converted into spatio-temporally invariant conditioning features $\bm{\Omega}_{1:T}$. The encoder $\mathcal{E}$ processes $\bm{\Omega}_{1:T}$ to predict a single body shape $\hat\beta$, a visibility context $\bm{\mathcal{G}}$, and per-frame summary features $\bm{\mathcal{S}}_{1:T}$. A diffusion decoder $\mathcal{D}$ then conditions on these features to denoise a noisy canonicalized pose $\mathbf{X}^n_{1:T}$ and, via global alignment, reconstruct the full-body motion $\mathbf{M}_{1:T}$.}
\label{fig:overview}
\end{figure}

\subsection{Problem Formulation}
Given an egocentric video from an arbitrary device, our goal is to reconstruct the wearer's 3D full-body motion.
Let $\mathbf{I}_{0:T}=\{I_t\}_{t=0}^T$ denote the egocentric image stream.
From $\mathbf{I}$, we compute head poses $\mathbf{T}^\text{head}_{0:T}=\{\mathbf{T}_t^\text{head}\in\mathrm{SE}(3)\}_{t=0}^T$ in world coordinates and intermittent wrist observations $\mathbf{H}_{0:T}=\{\mathbf{H}_t^\text{lHand},\mathbf{H}_t^\text{rHand}\}_{t=0}^T$.
Each $\mathbf{H}_t^\text{lHand/rHand}= (\mathbf{T}_t^\text{lHand/rHand},v_t^\text{lHand/rHand})$ consists of a 6-DoF wrist pose $\mathbf{T}_t^\text{lHand/rHand}\in\mathrm{SE}(3)$ in head coordinates and a visibility state $v_t^\text{lHand/rHand}\in\{0,1\}$.
A wrist may be invisible due to occlusion, motion blur, tracking failure, or moving outside the camera field of view.
Our model \modelname takes $\mathbf{T}^{\text{head}}_{0:T}$ and $\mathbf{H}_{0:T}$ as input and reconstructs human motion $\mathbf{M}_{1:T}$.
\begin{equation}
    \mathbf{M}_{1:T}=\modelname \left(\mathbf{T}^{\text{head}}_{0:T}, \mathbf{H}_{0:T} \right).
\end{equation}
The motion follows the SMPL-H format~\cite{romero2022embodied},
$\mathbf{M}_{1:T}=\{\beta, \bm{r}_{1:T}, \bm{\Phi}_{1:T}, \bm{\Theta}_{1:T}\}$,
where $\beta\in \mathbb{R}^{16}$ is a sequence-level shape, $\bm{r}_{1:T}$ is the root translation, $\bm{\Phi}_{1:T}$ is the root orientation, and $\bm{\Theta}_{1:T}$ denotes joint rotations.
The root poses are defined in the world coordinate system, consistent with $\mathbf{T}^{\text{head}}$.
Unlike prior approaches that use a fixed mean shape~\cite{lee2024mocapevery,jiang2022avatarposer,li2023egoego} or per-frame shape~\cite{yi2025egoallo,patel2025uniegomotion}, we predict a single consistent shape $\beta$ for the whole sequence.
Fig.~\ref{fig:overview} provides an overview of the framework.

\subsection{Preprocessing and Representation}
\label{sec:preprocessing_representation}

\paragraph{Preprocessing}
Given egocentric images $\mathbf{I}$, we first estimate camera poses $\mathbf{T}_t^\text{cam}\in\mathrm{SE}(3)$\footnote{$\mathbf{T}_t^\text{cam}$ denotes the transformation from the camera to world coordinates, equivalently $\mathbf{T}_t^{\text{cam} \to \text{world}}$. Similarly, $\mathbf{T}_t^\text{head} = \mathbf{T}_t^{\text{head} \to \text{world}}$.} using off-the-shelf SLAM.
We define a gravity-aligned world coordinate system with the floor at \(z=0\).
Since the camera and head joints are not co-located, we compute head poses as $\mathbf{T}_t^\text{head}=\mathbf{T}_t^\text{cam}(\mathbf{T}^{\text{cam}\to \text{head}})^{-1}$, where $\mathbf{T}^{\text{cam}\to \text{head}}$ is a mount-specific but time-invariant rigid transform estimated through lightweight pre-calibration.
We also extract 6-DoF wrist observations and visibility $\mathbf{H}_t^\text{hand}$ using off-the-shelf pose estimation modules.
For monocular RGB videos, we use HaWoR~\cite{zhang2025hawor}, without its infilling module, to extract both $\mathbf{T}_t^\text{cam}$ and $\mathbf{H}_t^\text{hand}$.
For Aria~\cite{engel2023projectarianewtool}, we use Aria Machine Perception Services (MPS).

\paragraph{Coordinate and Conditioning Representation}
Learning motion from long sparse sequences requires spatio-temporally invariant conditioning.
We therefore convert the raw signals $\left(\mathbf{T}^{\text{head}}_{t}, \mathbf{H}_t^\text{lHand}, \mathbf{H}_t^\text{rHand} \right)$ into a normalized representation.
For the head trajectory, following EgoAllo~\cite{yi2025egoallo}, we define a per-frame canonical frame $\mathbf{T}^{\text{cano}\to \text{world}}_t$ at the head's floor projection, with its \(z\)-axis aligned with gravity and \(y\)-axis aligned with the head's forward direction.
The condition vector $\bm{\Omega}_t \in \mathbb{R}^D$ is computed by a neural function $\Gamma$:
\begin{align}
\bm{\Omega}_t = \Gamma
\left(
\Delta \mathbf{T}^{\text{head}}_{t},\!
\mathbf{R}^{\text{head} \to \text{cano}}_{t},
h_t,
\Delta \mathbf{R}_t^{\text{cano}},\!
\mathbf{T}^{\text{hand} \to \text{head}}_{t}
\right).
\end{align}
Here, $\Delta \mathbf{T}^{\text{head}}_{t}$ is the relative head pose from frame $t-1$, $\mathbf{R}^{\text{head} \to \text{cano}}_{t}$ is the canonicalized head orientation, and $h_t$ is the head height from $\mathbf{T}_t^{\text{head}}$.
Unlike EgoAllo~\cite{yi2025egoallo}, we include the canonical relative rotation $\Delta \mathbf{R}_t^{\text{cano}}$ from frame $t-1$, because our model predicts the root orientation relative to the canonical frame.
We introduce hand conditions $\mathbf{T}^{\text{hand} \to \text{head}}_{t} = (\mathbf{T}^{\text{lHand} \to \text{head}}_{t},  \mathbf{T}^{\text{rHand} \to \text{head}}_{t})$, which encode relative hand poses in head coordinates.
When a hand is invisible, its cue is replaced with a learnable null embedding.
We do not include temporal hand motion cues, as wrist observations are often intermittent.

\paragraph{Global Alignment}
Following prior work~\cite{jiang2024egoposer,jiang2022avatarposer,yi2025egoallo}, we use $\mathbf{T}^\mathrm{head}$ for global localization and directly derive the root translation $\bm{r}_t$.
Unlike EgoAllo~\cite{yi2025egoallo}, however, we predict the relative root orientation $\mathbf{R}^{\text{root} \to \text{cano}}_{t}$ in the canonical coordinate.
The final global root orientation is computed as $\bm{\Phi}_t = \mathbf{R}^{\text{cano} \to \text{world}}_{t} \mathbf{R}^{\text{root} \to \text{cano}}_{t}$.
This representation provides an additional 3DoF in global orientation, improving motion stability and accuracy.

\subsection{Sequence-Level Diffusion Architecture}

We adopt a conditional diffusion model~\cite{ho2020denoisingdiffusionprobabilisticmodels} conditioned on the processed conditioning features $\bm{\Omega}_{1:T}= \{ \bm{\Omega}_t \}_{t=1}^T$.
The architecture separates \emph{sequence-level context} from local motion evidence.
A global pathway infers the wearer's body shape and camera-dependent visibility structure from long-range cues, while a local pathway summarizes per-frame head and hand evidence.
The two pathways are fused so that the same local evidence can be interpreted differently depending on the inferred body and observation context.
To train and infer on long sequences end-to-end without splitting them into independent subsequences, both the encoder $\mathcal{E}$ and decoder $\mathcal{D}$ use sliding-window local attention with horizon \(W\)~\cite{beltagy2020longformerlongdocumenttransformer, jiang2023mistral7b,barquero2024seamless,li2025genmo} and Rotary Positional Embedding~\cite{su2024roformer}, reducing attention complexity from quadratic to linear.

\paragraph{Sequence-Level Encoder}
The encoder $\mathcal{E}$ maps the sparse conditioning sequence $\bm{\Omega}_{1:T}$ into a sequence-level shape $\hat\beta$, a visibility context $\bm{\mathcal{G}}$, and per-frame summaries $\bm{\mathcal{S}}_{1:T}$.
\begin{equation}
\hat\beta = \mathcal{E}_\mathrm{shape}(\bm{\Omega}_{1:T}), \quad
\bm{\mathcal{G}} = \mathcal{E}_\mathrm{global}(\bm{\Omega}_{1:T}), \quad 
\bm{\mathcal{S}}_{1:T} = \mathcal{E}_\mathrm{local}(\bm{\Omega}_{1:T}, \bm{\mathcal{G}}).
\end{equation}
$\mathcal{E}_\mathrm{shape}$ infers the wearer's body proportions by pooling long-range cues such as head-height variations and the spatial extent of visible hand positions.
$\mathcal{E}_\mathrm{global}$ infers visibility context from visible hand poses and the patterns of when and where hands disappear.
Both branches first aggregate local temporal features with sliding-window attention and then pool information across the full sequence using learnable cross-attention queries~\cite{lee2019settransformerframeworkattentionbased}.
The local branch $\mathcal{E}_\mathrm{local}$ produces per-frame summaries from neighboring conditioning features, modulated by $\bm{\mathcal{G}}$ through AdaLN-Zero~\cite{Peebles2022DiT}.
This lets the same local head and hand evidence induce different poses depending on the inferred body shape and visibility context.

\paragraph{Decoder}
The decoder $\mathcal{D}$ denoises the canonicalized pose representation $\mathbf{X}_{1:T}^0=\{\mathbf{R}_{1:T}^{\text{root}\to\text{cano}},\bm{\Theta}_{1:T}\}$ rather than the final global motion directly.
The root translation $\bm{r}$ and orientation $\bm{\Phi}$ are then recovered as described in the Global Alignment section.
The decoder is implemented as a DiT~\cite{Peebles2022DiT} conditioned on the encoded summaries $\bm{\mathcal{S}}_{1:T}$ and body shape $\hat\beta$.
\begin{equation}
    \hat{\mathbf{X}}_{1:T}^0=
    \mathcal{D}(\mathbf{X}_{1:T}^n,n,\hat\beta,\bm{\mathcal{S}}_{1:T}).
\end{equation}
Here, $n\in\{1,\dots,N\}$ is the diffusion timestep and $\hat{\mathbf{X}}^0_{1:T}$ is the predicted $\mathbf{X}^0_{1:T}$.
The noised input $\mathbf{X}_{1:T}^n$ is defined by the DDPM~\cite{ho2020denoisingdiffusionprobabilisticmodels} forward process
$q(\mathbf{X}_{1:T}^n|\mathbf{X}_{1:T}^0)=\mathcal{N}(\mathbf{X}^n_{1:T};\sqrt{\bar{\alpha}_n}\mathbf{X}_{1:T}^0, (1-\bar\alpha_n)\mathbb{I})$, where $\bar\alpha_n$ is the noise schedule.
We inject global conditions $n$ and $\hat\beta$ via AdaLN, while the per-frame summary $\bm{\mathcal{S}}_{1:T}$ is injected via cross-attention using a local attention mask.

\subsection{Training}
\label{sec:augmentation}
\begin{figure}[t]\centering
\includegraphics[width=\linewidth, trim={0 0 0 0},clip]{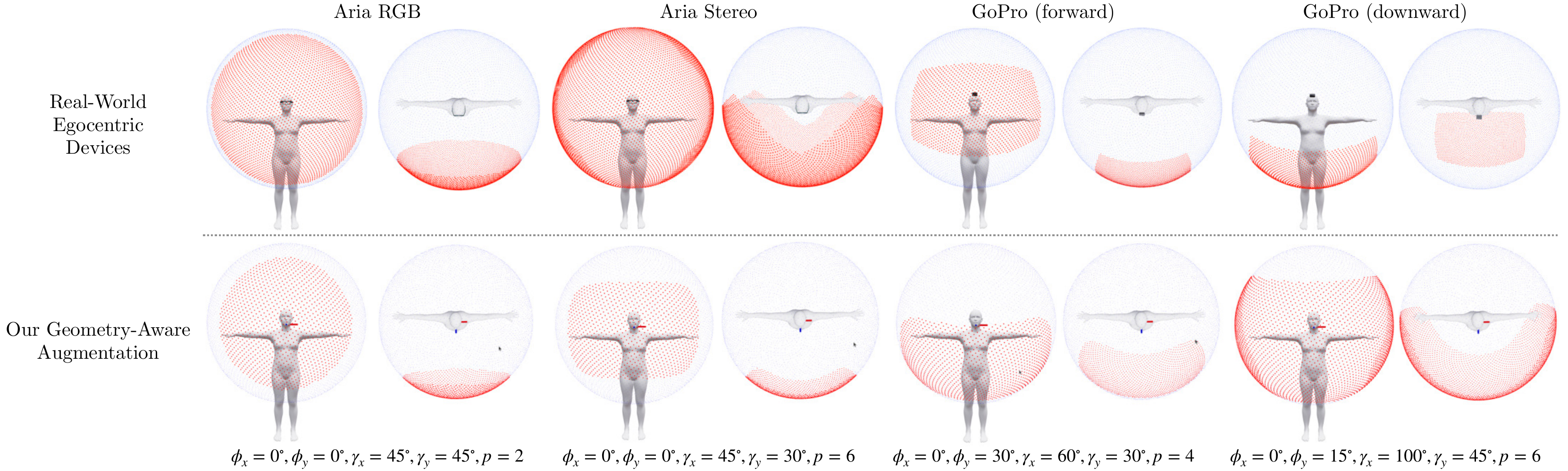}
\caption{\textbf{Real and simulated visibility boundaries.}
The top row shows calibrated FoVs from real-world devices (\eg, Aria, GoPro). 
The bottom row illustrates visibility patterns sampled by our geometry-aware augmentation.
Red regions indicate the visible field of view.}
\label{fig:augmentation}
\end{figure}

\paragraph{Geometry-Aware Visibility Augmentation}
For $\mathcal{E}_\mathrm{global}$ to learn visibility context rather than memorize a single visibility regime, training must expose the model to diverse camera-induced visibility patterns.
Motion-annotated egocentric datasets, such as Nymeria~\cite{ma2024nymeriamassivecollectionmultimodal}, obtain accurate motion using specialized capture hardware, but their hand visibility patterns are tied to specific camera optics and mountings.
Training directly on these patterns encourages a device-specific prior.
We instead synthesize a continuous family of camera-dependent visibility boundaries.

We model boundary variations caused by field of view, camera tilt, aspect ratio, lens masks, and optical distortions.
We rotate head coordinates by the center offsets $(\phi_x,\phi_y)$ into a camera frame, compute the wrist yaw $\psi_x$ and pitch $\psi_y$ in this frame, and define the boundary with half-angles $(\gamma_x,\gamma_y)$ and a power $p$ controlling boundary shape.
A wrist is visible when its angles satisfy
\begin{equation}
    \left|\psi_x\ /\ {\gamma_x}\right|^p+\left|\psi_y\ /\ {\gamma_y}\right|^p\le 1.
\end{equation}
Sampling these parameters from realistic ranges exposes the model to diverse camera setups, as shown in Fig.~\ref{fig:augmentation}.
Unlike random masking~\cite{aliakbarian2022flag}, this augmentation makes both visibility and invisibility depend on camera geometry, allowing $\mathcal{E}_\mathrm{global}$ to learn visibility patterns as sequence-level context.
We further handle imperfect in-the-wild tracking by applying stochastic temporal signal drops and Gaussian perturbations to visible wrist poses.
We sample signal drops from Poisson and Log-Normal distributions to model the heavy-tailed nature of occlusions.

\paragraph{Loss Functions}
Our training objective combines the DDPM simple loss, a shape consistency loss, and auxiliary physical losses.
\begin{equation}
    \mathcal{L}
    =\mathcal{L}_\mathrm{simple}
    +\lambda_\mathrm{shape}\mathcal{L}_\mathrm{shape}
    +\mathcal{L}_\mathrm{aux}.
\end{equation}
\begin{equation}
\mathcal{L}_\mathrm{simple}
=\mathbb{E}_{n,\mathbf{X}^0,\bm{\Omega}}
\left[\|\mathbf{X}^0-\hat{\mathbf{X}}^0 \|^2\right], \quad
\mathcal{L}_\mathrm{shape}
=\|\mathrm{FK}(\bm{0},\beta)-\mathrm{FK}(\bm{0},\hat{\beta})\|^2 .
\end{equation}
Here, $\mathcal{L}_\mathrm{simple}$ is the standard DDPM~\cite{ho2020denoisingdiffusionprobabilisticmodels} denoising objective.
To enforce shape consistency, $\mathcal{L}_\mathrm{shape}$ minimizes the 3D T-pose joint error instead of directly penalizing the PCA-derived $\beta$, which lacks physical meaning.
$\mathrm{FK}$ denotes the forward kinematics function.

Finally, $\mathcal{L}_\mathrm{aux}$ regularizes the reconstructed motion $\hat{\mathbf{M}}_{1:T}$ with joint position $\mathcal{L}_\mathrm{pos}$ and foot-skating $\mathcal{L}_\mathrm{skat}$ losses~\cite{tevet2023human}:
\begin{equation}
\mathcal{L}_\mathrm{pos}
\!\!=\!\!\frac{1}{T}\sum_{t=1}^T
\|\mathrm{FK}(\mathbf{M}_t,\beta)\!-\!\mathrm{FK}(\hat{\mathbf{M}}_t,\hat\beta)\|^2, \ 
\mathcal{L}_\mathrm{skat}
\!\!=\!\!\frac{1}{T\!\!-\!\!1}\sum_{t=1}^{T-1}
\|\mathrm{FK}(\hat{\mathbf{M}}_{t+1},\hat{\beta})\!
-\!\mathrm{FK}(\hat{\mathbf{M}}_t,\hat{\beta})\|^2\!\cdot c_t
\end{equation}
\begin{equation}
\mathcal{L}_\mathrm{aux}
=\bar{\alpha}_n
\left(\lambda_\mathrm{pos}\mathcal{L}_\mathrm{pos}
+\lambda_\mathrm{skat}\mathcal{L}_\mathrm{skat}\right).
\end{equation}
Here, $c_t$ is the ground-truth binary contact label, and scaling by $\bar{\alpha}_n$ emphasizes physical accuracy when the signal level is high.
We find that $\mathcal{L}_\mathrm{aux}$ is crucial for satisfying the hand condition $\mathbf{H}$ and generating stable and accurate motion.
We set $\lambda_\mathrm{shape}=2.0$, $\lambda_\mathrm{pos}=0.25$, and $\lambda_\mathrm{skat}=0.4$.

\subsection{Inference}
\label{sec:optimization}
We use DDIM~\cite{song2022denoisingdiffusionimplicitmodels} sampling for motion generation.
Because the denoiser is already conditioned on hand observations and sequence-level context, feed-forward prediction alone produces accurate and plausible motion without post-hoc fitting.
Optionally, we introduce timestep-adaptive guidance optimization that exploits the diffusion sampling process to improve visible wrist alignment while preserving motion plausibility.
We provide further details in the supplementary material.

\section{Experiments}
\label{sec:experiments}

\subsection{Experimental Setting}

\paragraph{Dataset}
We train on the AMASS~\cite{AMASS:ICCV:2019} train split with the proposed stochastic augmentation (Sec.~\ref{sec:augmentation}).
All sequences are resampled to 30 FPS and preprocessed following HuMoR~\cite{rempe2021humor}, including floor alignment and foot-contact annotation.
We evaluate on AMASS validation/test splits for simulation, our OmniEgoDB for real-world quantification, and in-the-wild videos for qualitative analysis.

\paragraph{Training Details}
We train with AdamW using a learning rate of $10^{-4}$, weight decay $10^{-4}$, and batch size 32 for 16 hours on 8 A5000 GPUs.
The maximum training sequence length is 512 frames, with attention horizon $W=\pm63$ implemented using FlexAttention~\cite{dong2024flexattentionprogrammingmodel}.
Test-time optimization uses the Theseus~\cite{pineda2022theseus} Levenberg--Marquardt optimizer.
Additional details are provided in the supplementary material.

\paragraph{Metrics}
We evaluate motion quality using Mean Per Joint Position Error (\textbf{MPJPE}, mm), Mean Per Joint Velocity Error (\textbf{MPJVE}, cm/s), and \textbf{Jerk} (km/s\textsuperscript{3}), the third derivative of position.
For hand reconstruction, we report Hand Position Error (\textbf{Hand PE}, mm) and Visible Hand Position Error (\textbf{Vis Hand PE}, mm), computed on visible hand frames.
For body shape, we report \textbf{Height Error} and \textbf{Span Error} (cm) for accuracy, and \textbf{Height Std} and \textbf{Span Std} (cm) for temporal consistency, where height and span are the maximum vertical and horizontal vertex distances in a T-pose.
\textbf{Runtime} (sec) measures average execution time per sequence on an RTX 3090Ti, including optimization setup. For the OmniEgoDB benchmark, SMPL-H predictions are evaluated against Xsens ground truth on 15 corresponding major joints, excluding ambiguous spine, collar, and toe joints.

\subsection{Overall Performance under Diverse Simulated Visibility}
\label{sec:result_diverse}

\paragraph{Setup}
We evaluate on simulated hand-visibility patterns generated by our augmentation and fixed across all methods.
We compare against EgoAllo~\cite{yi2025egoallo}, retrained on AMASS for a fair comparison under the same training distribution, with and without test-time optimization.
We also retrain EgoAllo$^\dagger$ and EgoPoser$^\dagger$~\cite{jiang2024egoposer} with our augmentation to isolate architectural differences, where EgoAllo$^\dagger$ denotes our hand-conditioned EgoAllo extension.
For shape metrics, baselines use mean shape for spatial errors and raw per-frame estimates for consistency.

\begin{table}[!t]
\centering
\caption{
\textbf{Quantitative results under diverse simulated settings.} w/o opt denotes evaluation without optimization; w/ opt denotes evaluation with optimization. 
}
\label{table:quant_diverse}
\setlength{\aboverulesep}{0pt}
\setlength{\belowrulesep}{0pt}
\renewcommand{\arraystretch}{1.47}
\setlength{\tabcolsep}{4.7pt}
\resizebox{\textwidth}{!}{
\begin{NiceTabular}{l|ccc|cc|cccc|c}
\CodeBefore
  \rowcolor{gray!15}{3,6,8}
\Body
\toprule
\rule[-5pt]{0pt}{16pt}Method & MPJPE${\downarrow}$ & MPJVE${\downarrow}$ & Jerk${\downarrow}$ & Hand PE${\downarrow}$ & Vis Hand PE${\downarrow}$ & Height${\downarrow}$ & Span${\downarrow}$ & Height Std${\downarrow}$ & Span Std${\downarrow}$ & Runtime${\downarrow}$ \\ 
\midrule
EgoAllo (w/o opt) & 122.99 & 44.02 & 0.350 & 318.69 & 298.01 & 3.80 & 6.24 & 1.87 & 2.43 & 2.49 \\
EgoAllo (w/ opt) & 99.32 & 31.66 & 0.152 & 164.80 & 41.49 & 3.86 & 6.19 & 1.84 & 2.35 & 56.09 \\
\midrule
EgoPoser$^{\dagger}$ & 117.04 & 58.24 & 0.972 & 244.39 & 117.45 & 4.65 & 7.52 & 2.97 & 3.52 & \textbf{0.09} \\
EgoAllo$^{\dagger}$ (w/o opt) & 103.55 & 35.57 & 0.290 & 212.95 & 122.89 & 3.62 & 6.10 & 1.70 & 2.06 & 2.55 \\
EgoAllo$^{\dagger}$ (w/ opt) & 94.00 & 26.13 & 0.120 & 142.24 & \textbf{25.74} & 3.78 & 6.25 & 1.71 & 2.08 & 59.26 \\
\midrule
Ours (w/o opt) & 81.35 & 25.81 & \textbf{0.117} & 146.71 & 56.98 & \textbf{2.40} & \textbf{4.61} & \textbf{0.00} & \textbf{0.00} & 2.07 \\
Ours (w/ opt) & \textbf{80.45} & \textbf{25.48} & 0.120 & \textbf{132.74} & 29.34 & \textbf{2.40} & \textbf{4.61} & \textbf{0.00} & \textbf{0.00} & 5.97
\\
\bottomrule
\end{NiceTabular}
}
\end{table}

\begin{figure*}[t]\centering
\includegraphics[width=\linewidth, trim={0 0 0 0},clip]{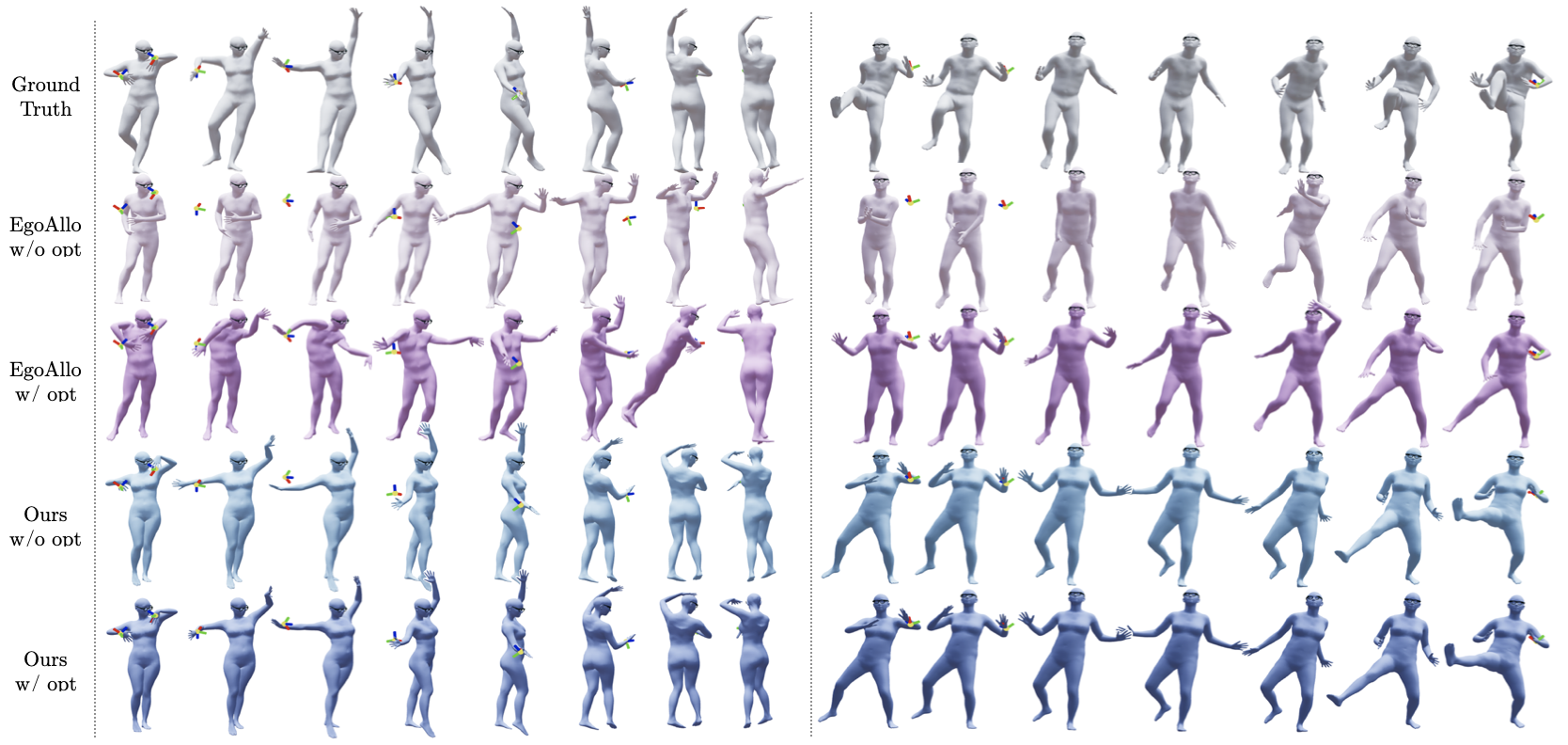}
\caption{\textbf{Qualitative results under diverse simulated settings.}
\textbf{(Left)} EgoAllo produces unnatural, unstable motion during dancing, while ours remains plausible.
\textbf{(Right)} During a kick, EgoAllo hallucinates the invisible hand back into the FoV, whereas ours predicts it outside the view.
}
\label{fig:qual_diverse}
\end{figure*}

\paragraph{Results}
Table~\ref{table:quant_diverse} shows that our model achieves the best overall motion, hand, and shape metrics while predicting a single consistent body shape across the sequence.
This supports our sequence-level formulation: body shape helps disambiguate full-body motion from head trajectories, while visibility context helps interpret intermittent hand observations.
Although EgoAllo$^\dagger$ with optimization attains the lowest visible-hand error, its higher overall Hand PE reveals failures during invisible periods.
As shown in Fig.~\ref{fig:qual_diverse}, EgoAllo often hallucinates unobserved hands inside the FoV and then snaps them to the observations when they reappear.
Its aggressive post-hoc optimization can also introduce artifacts such as body penetration or implausible joint angles when enforcing sparse hand constraints.
In contrast, by conditioning on intermittent hands and sequence-level visibility context, ours keeps out-of-FoV hand trajectories plausible and avoids such discontinuities.
Moreover, baselines with per-frame shape estimates can change body proportions even under the same motion state, amplifying ambiguities such as a shorter wearer standing versus a taller wearer sitting, whereas our sequence-level shape remains fixed.
Our lightweight guidance further improves hand alignment while preserving smooth motion and running over $9{\times}$ faster than EgoAllo.
Finally, our improvement over EgoPoser$^\dagger$ and EgoAllo$^\dagger$ shows that augmented training alone is insufficient without our sequence-level architecture.

\subsection{Generalization across Camera Setups}
\label{sec:result_fixed}

\paragraph{Setup}
We evaluate robustness to camera-geometry shifts using five synthetic visibility setups: $90^\circ$, $120^\circ$, and $180^\circ$ FoV, $120^\circ$ FoV with $30^\circ$ downward tilt, and $120^\circ$ hFoV $\times$ $60^\circ$ vFoV.
To isolate geometry effects, all models are trained and tested without temporal drops or pose perturbations.
Baselines marked (120) are trained under EgoPoser's fixed $120^\circ$ pinhole visibility regime, whereas \modelname uses only geometry-aware visibility augmentation.
EgoAllo$^{\dagger\dagger}$ is our hand-conditioned EgoAllo extension trained under this fixed $120^\circ$ regime.
Table~\ref{table:quant_fixed} also includes visibility-modeling variants of \modelname, analyzed in Sec.~\ref{sec:ablation}.

\begin{table}[!t]
\centering
\caption{\textbf{Quantitative results across varying camera setups.} PE and H-PE denote MPJPE and Hand PE.}
\label{table:quant_fixed}
\scriptsize
\setlength{\tabcolsep}{2.2pt}
\setlength{\aboverulesep}{0pt}
\setlength{\belowrulesep}{0pt}
\renewcommand{\arraystretch}{1.4}
\resizebox{\textwidth}{!}{
\begin{NiceTabular}{l | ccc | ccc | ccc | ccc | ccc }
\CodeBefore
  \rowcolor{gray!15}{5,9}
\Body
\toprule
\multirow{2}{*}{Method} & \multicolumn{3}{c|}{$90^\circ$ FoV} & \multicolumn{3}{c|}{$120^\circ$ FoV} & \multicolumn{3}{c|}{$180^\circ$ FoV} & \multicolumn{3}{c|}{$120^\circ$ + $30^\circ$ Tilt} & \multicolumn{3}{c}{$120^\circ \times 60^\circ$} \\
\cmidrule{2-16}
& PE$\downarrow$ & Jerk$\downarrow$ & H-PE$\downarrow$ & PE$\downarrow$ & Jerk$\downarrow$ & H-PE$\downarrow$ & PE$\downarrow$ & Jerk$\downarrow$ & H-PE$\downarrow$ & PE$\downarrow$ & Jerk$\downarrow$ & H-PE$\downarrow$ & PE$\downarrow$ & Jerk$\downarrow$ & H-PE$\downarrow$ \\
\midrule
EgoPoser (120) & 100.80 & 0.519 & \textbf{164.19} & 95.62 & 0.505 & \textbf{136.04} & 88.95 & 0.427 & 100.93 & 98.02 & 0.526 & 150.97 & 104.63 & 0.525 & \textbf{183.79} \\
EgoAllo$^{\dagger\dagger}$ (120) & 110.23 & 0.298 & 212.93 & 107.10 & 0.293 & 178.94 & 113.14 & 0.284 & 156.06 & 115.86 & 0.287 & 183.02 & 113.94 & 0.299 & 237.91 \\
\midrule
EgoAllo (w/ opt) & 112.06 & 0.162 & 239.87 & 104.80 & 0.165 & 192.48 & 86.70 & 0.159 & 75.56 & 91.36 & 0.156 & 108.28 & 115.13 & 0.161 & 259.06 \\
\midrule
Ours (120) & 101.56 & 0.129 & 207.62 & 91.11 & 0.126 & 152.14 & 92.07 & 0.115 & 130.77 & 95.64 & 0.117 & 160.50 & 111.58 & 0.131 & 266.18 \\
Ours (Random) & 151.73 & 0.158 & 278.73 & 131.97 & 0.154 & 246.58 & 94.78 & 0.139 & 137.02 & 103.51 & 0.142 & 172.89 & 161.32 & 0.159 & 295.27 \\
\midrule
Ours & 93.65 & \textbf{0.119} & 199.35 & 85.70 & \textbf{0.117} & 158.33 & 72.22 & \textbf{0.114} & 88.50 & 75.70 & \textbf{0.115} & 110.64 & 98.00 & \textbf{0.119} & 223.59 \\
Ours (w/ opt) & \textbf{93.21} & 0.120 & 193.40 & \textbf{85.32} & 0.119 & 149.65 & \textbf{71.67} & 0.117 & \textbf{70.96} & \textbf{75.36} & 0.117 & \textbf{96.23} & \textbf{97.48} & 0.120 & 217.47 \\
\bottomrule 
\end{NiceTabular}
}
\end{table}

\paragraph{Results}
Table~\ref{table:quant_fixed} shows that our model generalizes consistently across camera geometries, achieving the best full-body accuracy and smoothness across all five setups.
The fixed-regime baselines reveal two complementary failure modes.
EgoPoser (120) can obtain low Hand PE by collapsing invisible hands toward a conservative mean pose, but this conservative strategy leads to higher full-body error, jerk, and discontinuities near FoV transitions.
Although EgoAllo$^{\dagger\dagger}$ uses a diffusion prior, it overfits to the fixed $120^\circ$ visibility boundary and even worsens at $180^\circ$ despite increased visibility, showing that diffusion priors alone do not ensure camera-geometry generalization.
Supplementary Fig.~\ref{fig:qual_fixed} shows this boundary bias, where EgoAllo$^{\dagger\dagger}$ pushes invisible hands toward its learned boundary and fails to adapt to the actual FoV.
Ours also outperforms DSPoser~\cite{chi2024estimatingegobodyposedoubly} in its own $90^\circ$ setting (Supplementary Table~\ref{tab:dsposer}).
Overall, geometry-aware augmentation and sequence-level visibility context allow ours to adapt to unseen FoVs and mountings without device-specific retraining while preserving plausible full-body motion.

\subsection{Real-World Evaluation on OmniEgoDB and Further In-the-Wild Results}
\label{sec:real_world}

\begin{table}[t]
\centering
\caption{
\textbf{Quantitative results on OmniEgoDB.}
PE, VE, and H-PE denote MPJPE, MPJVE, and Hand PE.
$^\#$ denotes methods evaluated with their official checkpoints.
-- denotes entries not applicable to UniEgoMotion, which needs RGB frames and predicts at 10 FPS.
}
\label{table:quant_benchmark}
\scriptsize
\setlength{\tabcolsep}{2.2pt}
\setlength{\aboverulesep}{0pt}
\setlength{\belowrulesep}{0pt}
\renewcommand{\arraystretch}{1.4}
\resizebox{\textwidth}{!}{
\begin{NiceTabular}{l|cccc|cccc|cccc|cccc}
\toprule
\multirow{3}{*}{Method}
& \multicolumn{4}{c|}{Aria (Stereo)}
& \multicolumn{4}{c|}{Aria (RGB)}
& \multicolumn{4}{c|}{GoPro}
& \multicolumn{4}{c}{Quest} \\[-3.5pt]
& \multicolumn{4}{c|}{\tiny 21 seqs, 67.1 min}
& \multicolumn{4}{c|}{\tiny 21 seqs, 67.1 min}
& \multicolumn{4}{c|}{\tiny 20 seqs, 68.1 min}
& \multicolumn{4}{c}{\tiny 9 seqs, 42.0 min} \\
\cmidrule{2-17}
& PE$\downarrow$ & VE$\downarrow$ & Jerk$\downarrow$ & H-PE$\downarrow$
& PE$\downarrow$ & VE$\downarrow$ & Jerk$\downarrow$ & H-PE$\downarrow$
& PE$\downarrow$ & VE$\downarrow$ & Jerk$\downarrow$ & H-PE$\downarrow$
& PE$\downarrow$ & VE$\downarrow$ & Jerk$\downarrow$ & H-PE$\downarrow$ \\
\midrule
EgoAllo$^\#$ (w/o opt)
& 138.22 & 24.70 & 0.146 & 261.10
& 138.22 & 24.70 & 0.146 & 261.10
& 136.86 & 24.11 & 0.143 & 260.80
& 152.17 & 22.02 & 0.140 & 282.93 \\
EgoAllo$^\#$ (w/ opt)
& 115.48 & 25.06 & 0.122 & 142.00
& 128.71 & 32.26 & 0.259 & 182.86
& 125.65 & 28.65 & 0.189 & 186.80
& 122.87 & 22.97 & 0.121 & 135.15 \\
EgoPoser$^\#$
& 126.54 & 43.56 & 0.659 & 172.75
& 135.99 & 51.65 & 0.906 & 197.39
& 134.89 & 47.30 & 0.776 & 209.21
& 130.04 & 35.17 & 0.521 & 153.07 \\
EgoPoser (120)
& 124.95 & 58.15 & 1.265 & 185.71
& 132.22 & 65.00 & 1.551 & 198.19
& 130.40 & 65.67 & 1.580 & 213.15
& 125.46 & 46.07 & 0.979 & 161.43 \\
UniEgoMotion$^\#$
& -- & -- & -- & --
& 151.61 & -- & -- & 243.02
& 161.10 & -- & -- & 258.63
& -- & -- & -- & -- \\
\midrule
Ours
& 106.61 & \textbf{21.29} & \textbf{0.077} & 142.28
& \textbf{118.11} & \textbf{23.25} & \textbf{0.083} & 173.83
& \textbf{111.68} & \textbf{21.81} & \textbf{0.073} & 166.20
& \textbf{110.49} & \textbf{18.16} & \textbf{0.062} & 135.86 \\
Ours (w/ opt)
& \textbf{106.23} & 21.81 & 0.093 & \textbf{134.56}
& 119.12 & 25.99 & 0.178 & \textbf{171.28}
& 112.21 & 23.30 & 0.125 & \textbf{164.39}
& 111.50 & 18.91 & 0.080 & \textbf{128.35} \\
\bottomrule
\end{NiceTabular}
}
\end{table}

\begin{figure}[t]\centering
\includegraphics[width=\linewidth, trim={0 0 0 0},clip]{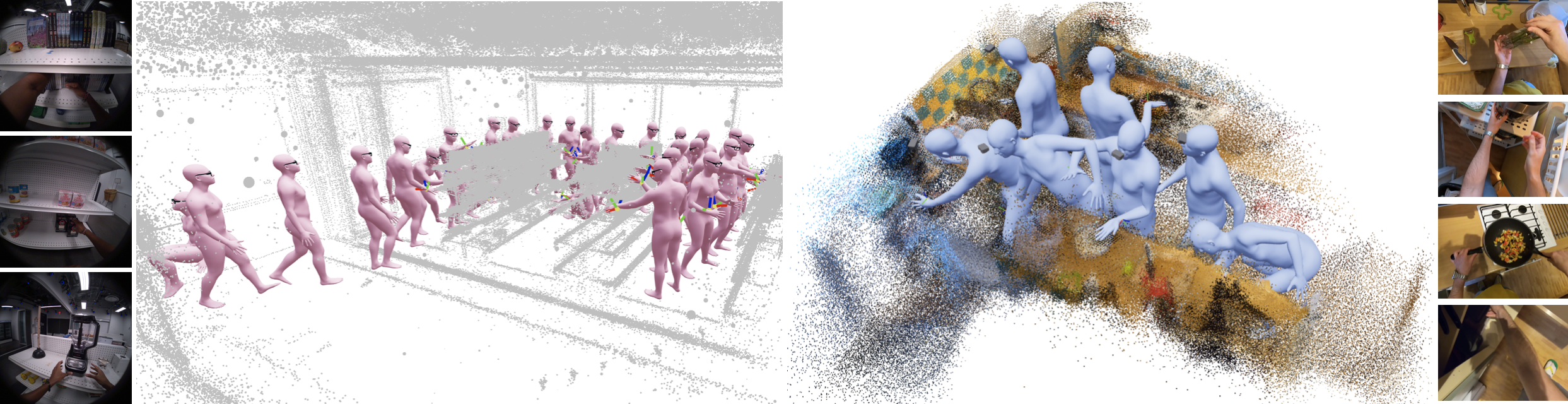}
\caption{\textbf{Qualitative results on in-the-wild data.} Reconstructions from (Left) Aria RGB (monocular) and (Right) GoPro captures.}
\label{fig:in_the_wild}
\end{figure}

Existing egocentric motion datasets are limited for evaluating device-agnostic reconstruction, as they are typically captured with a single device, most commonly Project Aria glasses~\cite{metaaria}, and rely on specialized hardware whose mountings induce device-specific visibility patterns~\cite{ma2024nymeriamassivecollectionmultimodal}. We introduce \textbf{OmniEgoDB}, a new mocap benchmark that pairs Xsens bodysuit ground-truth motion with recordings from multiple consumer egocentric devices, including Aria, GoPro, and Quest, across 50 diverse sequences from four subjects totaling 177 minutes. This enables direct evaluation of cross-device generalization under real-world conditions.
Capture details are provided in the supplementary material.

\begin{figure}[t]
    \centering
    \begin{minipage}{0.57\textwidth}
        \centering
        \includegraphics[width=\linewidth]{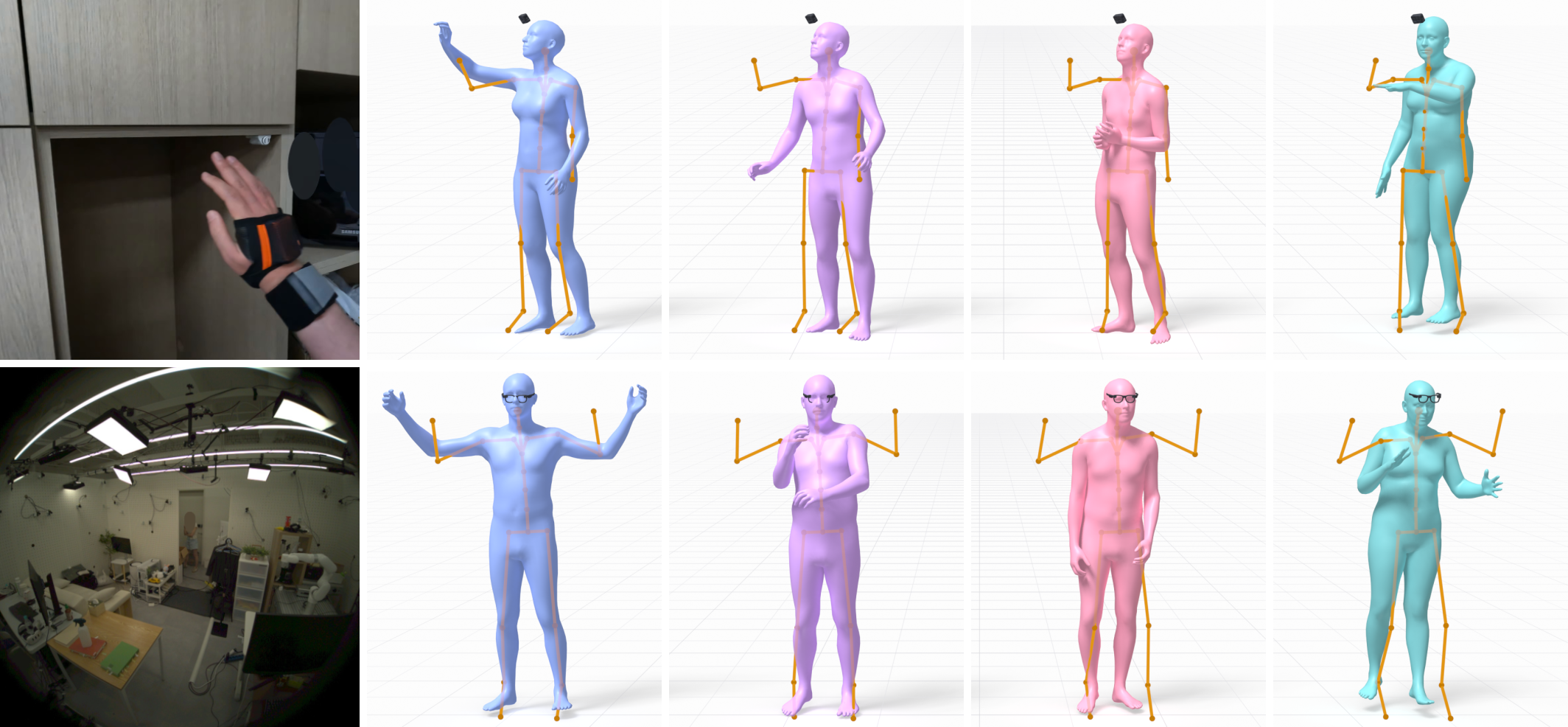}
        \captionof{figure}{\textbf{Qualitative results on OmniEgoDB.} \mbox{GoPro} (top) and Aria (RGB) (bottom). From left, input, \mbox{\modelname}, EgoAllo$^\#$ (w/ opt), EgoPoser (120), and UniEgoMotion$^\#$, with the Xsens ground truth in orange.}
        \label{fig:qual_omniegodb_main}
    \end{minipage}
    \hfill
    \begin{minipage}{0.40\textwidth}
        \resizebox{\linewidth}{!}{
            \input{figures/07_qual_egoexo/figure}
        }
        \captionof{figure}{\textbf{Qualitative results on EgoExo4D.} From top, input, EgoAllo (w/ opt), and ours (w/ opt). EgoAllo's shape inconsistency worsens under real-world noise, causing fluctuations during static cooking, while ours remains stable throughout.
        }
        \label{fig:qual_egoexo}
    \end{minipage}
\end{figure}

Table~\ref{table:quant_benchmark} reports quantitative results on OmniEgoDB.
Aria (Stereo) uses the device's hand tracking, and Aria (RGB) uses HaWoR~\cite{zhang2025hawor} on the RGB stream of the same recordings.
We run the official EgoPoser~\cite{jiang2024egoposer} at its native 60 FPS on inputs upsampled from 30 FPS and the official UniEgoMotion~\cite{patel2025uniegomotion} at its native 10 FPS, and resample their predictions to 30 FPS for evaluation.
The official EgoPoser does not estimate body shape, so we evaluate it with the mean shape.
Unlike simulated settings, OmniEgoDB contains real-world device recordings with longer trajectories and larger spatial coverage, while also introducing diverse optics and mountings, imperfect wrist visibility estimates, and noisy 6-DoF wrist pose estimates.
Across all devices and both Aria settings, \modelname achieves the lowest MPJPE, MPJVE, and Jerk, and with guidance the lowest Hand PE, showing that the learned body shape and visibility context transfer to real-world device recordings without device-specific retraining (Fig.~\ref{fig:qual_omniegodb_main}).
No method, including ours, is trained on OmniEgoDB.
We also report zero-shot results on Aria Digital Twin~\cite{pan2023ariadigitaltwin} in the supplementary material.

Beyond OmniEgoDB, we evaluate \modelname on public in-the-wild egocentric videos from EgoExo4D~\cite{grauman2024egoexo4dunderstandingskilledhuman}, Reading in the Wild~\cite{yang25reading}, EPIC-KITCHENS~\cite{Damen2022RESCALING}, and custom captures.
As shown in Fig.~\ref{fig:teaser}, Fig.~\ref{fig:qual_egoexo}, and Fig.~\ref{fig:in_the_wild}, \modelname reconstructs coherent full-body motion across diverse camera configurations without device-specific retraining.
Baselines often exhibit jitter, abrupt pose changes, or implausible arm motion in these noisy settings, while ours maintains stable global motion and consistent body shape.
This supports real-world deployment from ordinary egocentric videos.

\subsection{Ablation Studies}
\label{sec:ablation}

\begin{figure}[t]
    \centering
    \begin{minipage}{0.50\textwidth}
        \captionof{table}{\textbf{Quantitative results on ablation studies.} Comparison of variants on diverse simulated hand observations. All variants are evaluated without optimization.}
        \label{table:ablation}
        \centering
        \resizebox{\linewidth}{!}{
            \footnotesize 
\setlength{\tabcolsep}{2pt} %
\renewcommand{\arraystretch}{1.41}
\begin{NiceTabular}{l|ccc|cc}
\toprule
Method & MPJPE${\downarrow}$ & MPJVE${\downarrow}$ & Jerk${\downarrow}$ & Hand PE${\downarrow}$ & Vis Hand PE${\downarrow}$ \\ 
\midrule
No Shape/Global & 97.06 & \textbf{25.17} & \textbf{0.104} & 161.25 & 79.70 \\
No Shape Cond & 86.55 & 26.59 & 0.111 & 156.09 & 65.05 \\
No Global & 83.53 & 26.82 & 0.116 & 152.73 & 65.99 \\
Concat & 81.60 & 26.13 & 0.112 & 149.92 & 61.42 \\
\midrule
Default & \textbf{81.35} & 25.81 & 0.117 & \textbf{146.71} & \textbf{56.98} \\
\bottomrule
\end{NiceTabular}

        }
        
    \end{minipage}
    \hfill
    \begin{minipage}{0.47\textwidth}
        \centering
        \captionof{table}{\textbf{Linear probing of the visibility context.} Visible-region IoU (top) and stature error in cm (bottom) on held-out AMASS motions.}
        \label{table:probing}
        \centering\scriptsize
        \setlength{\tabcolsep}{4pt}
        \setlength{\aboverulesep}{0pt}
        \setlength{\belowrulesep}{0pt}
        \renewcommand{\arraystretch}{1.348}
        \begin{NiceTabular}{X[l]|c}[width=\linewidth]
        \toprule
        Probe input (frozen) & Region IoU$\uparrow$ \\
        \midrule
        Shuffled labels & 0.447 \\
        Mean camera & 0.582 \\
        Visibility statistics & 0.662 \\
        \midrule
        $\bm{\mathcal{G}}$ (ours) & \textbf{0.740} \\
        \midrule
        Probe input (frozen) & Stature err.$\downarrow$ \\
        \midrule
        $\bm{\mathcal{G}}$ & 9.19 \\
        $\hat\beta$ & \textbf{3.05} \\
        \bottomrule
        \end{NiceTabular}
    \end{minipage}
\end{figure}

We conduct ablation studies to validate the two components central to our formulation: geometry-aware visibility augmentation and sequence-level context.
Additional ablations on the auxiliary loss and canonical root-orientation representation are provided in the supplementary material.

\paragraph{Necessity of Geometry-Aware Augmentation}
Table~\ref{table:quant_fixed} evaluates two visibility-modeling variants: \textbf{Ours (Random)}, which uses random masking with probability 0.2~\cite{aliakbarian2022flag,jiang2024egoposer}, and \textbf{Ours (120)}, trained with a fixed $120^\circ$ visibility boundary.
Random masking performs poorly because it removes hand observations without preserving camera-induced spatial structure.
The fixed-boundary variant behaves like a camera-specific specialist, performing competitively near its native setup but degrading under out-of-distribution FoVs and mountings.
In contrast, our geometry-aware augmentation exposes the model to a continuous family of visibility boundaries, enabling it to learn transferable visibility context. Notably, geometry-aware augmentation improves full-body accuracy over the fixed-boundary variant even at the nominal $120^\circ$ setting, while maintaining comparable hand accuracy.

\paragraph{Importance of Sequence-Level Context}
Table~\ref{table:ablation} ablates our sequence-level shape and visibility context.
\textbf{No Shape/Global} removes both branches and uses the mean shape without $\bm{\mathcal{G}}$.
\textbf{No Shape Cond} predicts $\hat{\beta}$ but removes decoder shape conditioning.
\textbf{No Global} removes $\bm{\mathcal{G}}$, and \textbf{Concat} replaces AdaLN with concatenation.
Removing all sequence-level context (No Shape/Global) substantially degrades MPJPE, while shape conditioning alone (No Global) recovers most full-body accuracy and adding $\bm{\mathcal{G}}$ (Default) further improves hand accuracy.
Supplementary qualitative results support this interpretation, showing that $\bm{\mathcal{G}}$ captures camera-induced visibility boundaries and helps maintain FoV-consistent invisible-hand trajectories.
These results demonstrate that body shape and visibility context are most effective when used to modulate the denoising process, rather than being injected as local conditioning alone.

\paragraph{Probing the Visibility Context}
To test whether $\bm{\mathcal{G}}$ encodes camera geometry, we freeze the encoder and fit a ridge-regression probe from $\bm{\mathcal{G}}$ to the camera's visible region on a sphere of head-relative directions, using held-out AMASS test motions each paired with 12 cameras sampled from our augmentation.
The IoU between predicted and true regions is computed over the directions the wrists visit, with folds grouped by motion (details in the supplementary material).
$\bm{\mathcal{G}}$ reaches 0.740 IoU (Table~\ref{table:probing}), well above shuffled labels (0.447), the mean training camera (0.582), and coarse per-sequence visibility statistics such as the fraction of frames with visible wrists and the visibility transition rate (0.662), so it encodes more than how often the wrists are seen.
In contrast, $\hat\beta$ from the same encoder predicts wearer stature (3.05~cm error, versus 9.19~cm from $\bm{\mathcal{G}}$), indicating that the two branches separate camera geometry from body shape.

\section{Discussion}

We present \textbf{\modelname}, a sequence-level diffusion framework for 3D full-body motion reconstruction across diverse egocentric camera setups.
By inferring consistent body shape and visibility context from long-range cues, \modelname resolves head-trajectory ambiguity and treats intermittent hand visibility as geometric evidence rather than missing data.
Together with geometry-aware visibility augmentation, this enables device-agnostic reconstruction without device-specific retraining.
We also introduce \textbf{OmniEgoDB}, a multi-device mocap benchmark for quantitative cross-device evaluation with real-world egocentric recordings and ground-truth full-body motion.

Our framework needs no device-specific retraining, but each device still requires calibration and preprocessing that provide a mount-specific camera-to-head transform, a gravity-aligned metric head trajectory, and wrist observations with visibility flags.
Devices without native tracking obtain these from SLAM and off-the-shelf hand estimators, so reconstruction quality depends on these upstream components.
\modelname is designed for offline sequence-level reconstruction, not low-latency real-time inference.
It also assumes a flat ground plane and does not model finger articulation or hand-object interactions.
OmniEgoDB covers only four subjects and a limited range of capture locations and activities.
Future work may improve robustness to tracking failures, incorporate non-planar scene geometry, and model finger articulation and richer hand-object interactions for more general egocentric motion understanding.

\clearpage

\begin{ack}
We thank Gunhee Kim and Lee Chae-Yeon for their help with the benchmark capture.
This work was supported by KT (Korea Telecom)-Seoul National University AICT R\&D Center. This work was also supported by NRF grant funded by the Korean government (MSIT) (No. RS-2022-NR070498), IITP grant funded by the Korea government (MSIT) [No. RS-2024-00439854, No. 2022-0-00156, and No. RS-2025-25441838], and Institute of Engineering Research at Seoul National University. H. Joo is the corresponding author.
\end{ack}

{
\small
\bibliographystyle{plain}
\bibliography{sections/06_references}
}

\appendix
\clearpage

\section{Implementation Details}
\label{sec:supp_impl_details}

In this section, we provide additional details to supplement the descriptions in the main text. Code, pre-trained checkpoints, and OmniEgoDB are publicly available on our project page.

\paragraph{Network Architecture}
Tab.~\ref{tab:architecture} summarizes the capacity of each sub-network. While the core structural design is introduced in the main text, we detail the specific module configurations here. Prior to the transformer blocks, raw condition variables are projected into their respective hidden dimensions via Multi-Layer Perceptrons (MLPs). For the shape and global branches, the attention poolers~\cite{lee2019settransformerframeworkattentionbased} use 16 learnable queries that attend to the input sequence, are refined by self-attention, then mean-pooled and passed through an MLP to produce the branch output, \ie $\hat{\beta}$ for the shape branch and $\bm{\mathcal{G}}$ for the global branch. Throughout the network, all transformer blocks adopt pre-Layer Normalization and GELU activations. In the decoder, the global conditioning vector for the AdaLN-Zero~\cite{Peebles2022DiT} modules is constructed by directly adding the Fourier-embedded diffusion timestep $n$ and the MLP-projected shape parameter $\hat{\beta}$.

\paragraph{Training and Inference}
We apply an Exponential Moving Average (EMA) decay of 0.9999 to stabilize training. For the diffusion process, our model is parameterized to directly predict the clean canonicalized pose $\mathbf{X}^0$ ($\mathbf{X}_0$-prediction). We utilize a cosine noise schedule with 1,000 DDPM~\cite{ho2020denoisingdiffusionprobabilisticmodels} training timesteps, while inference is accelerated using 30 DDIM~\cite{song2022denoisingdiffusionimplicitmodels} steps.
For the test-time guidance optimization introduced in the main text, we specifically optimize only the 3D local rotations of the arm joints (\eg, shoulder, elbow, wrist). To ensure computational efficiency, the Levenberg-Marquardt solver is constrained to a maximum of 5 iterations per diffusion step and is executed in a fully vectorized manner.

\section{Augmentation Details} We present the detailed hyperparameters for the augmentation strategy introduced in Sec.~\ref{sec:augmentation}. 

\paragraph{Geometry-Aware Augmentation} 
To simulate diverse capture setups, ranging from monocular and stereo to fisheye systems with varying orientations (forward to downward), we sample spatial parameters uniformly. The sampling ranges are defined as follows (angles in radians):
\begin{equation}
    \begin{gathered}
        \gamma_x \in [0.35, 2.15], ~ \gamma_y \in [0.35, 1.35] \\
        \phi_x \in [-0.15, 0.15], ~ \phi_y \in [0.0, 1.5], ~ p \in [2.0, 10.0] 
    \end{gathered}
\end{equation}
To preserve realistic aspect ratios and tilt orientations, we enforce the constraints:
\begin{equation}
    0.4 \le \frac{\gamma_y}{\gamma_x} \le 1.1, \quad \phi_y \le 1.5 - (\gamma_y - 0.35) \cdot 1.1.
\end{equation}

\paragraph{Temporal Drops and Pose Perturbation} 
To simulate realistic tracking failures like motion blur or prolonged occlusions, we employ a two-stage stochastic masking strategy. We dynamically sample the number of drop events $K \sim \text{Poisson}(T \cdot \rho / \mathbb{E}[D])$ for a sequence of length $T$ to match a target drop ratio $\rho$. The duration $D$ of each drop is drawn from a heavy-tailed Log-Normal distribution with mean $m=\mathbb{E}[D]$ and standard deviation $s$. We define two independent masking modes:
\begin{equation}
\begin{split}
    \text{Short:} ~ & \rho \sim \mathcal{U}(0.0, 0.1), ~ m=2.0, ~ s=1.0 \\
    \text{Long:} ~ & \rho \sim \mathcal{U}(0.0, 0.2), ~ m=28.0, ~ s=25.0
\end{split}
\end{equation}
The final visibility mask is the union of these two modes, clipping the long-drop durations to a minimum of 5 frames.

Finally, to mimic systematic errors of off-the-shelf wrist estimators, we offset the wrist poses of each hand by a small random rigid transformation in head coordinates, sampled once per sequence, with rotation angles drawn from $\mathcal{N}(0, (\pi/180)^2)$ along each axis (in radians) and translation drawn from $\mathcal{N}(0, 0.005^2)$ (in meters).

\begin{table}[t]
    \centering
    \caption{\textbf{Architectural hyperparameters.} Configurations of the transformer modules in \modelname.}
    \label{tab:architecture}
    \scriptsize 
    \setlength{\tabcolsep}{5pt}
\setlength{\aboverulesep}{0pt}
\setlength{\belowrulesep}{0pt}
\renewcommand{\arraystretch}{1.4}
\begin{NiceTabular}{X[3,l]|X[2,c]X[2,c]X[2,c]X[2,c]X[2,c]}[width=\linewidth]
\toprule
Module & Hidden Dim & Out Dim & Layers & Heads & Dropout \\
\midrule
Shape Encoder & 384 & 384 & 2 & 6 & 0.1 \\
Shape Pooler & 384 & 16 & 1 & 6 & 0.1 \\
Global Encoder & 384 & 384 & 2 & 6 & 0.1 \\
Global Pooler & 384 & 512 & 1 & 6 & 0.1 \\
Local Encoder & 512 & 512 & 6 & 8 & 0.1 \\
Decoder & 512 & 132 & 8 & 8 & 0.1 \\
\bottomrule
\end{NiceTabular}

\end{table}

\begin{table}[t]
    \centering
    \begin{minipage}[t]{0.53\textwidth}
        \centering
        \caption{\textbf{Additional ablation studies.} Auxiliary loss and root-orientation representation on diverse simulated hand observations. All variants are evaluated without optimization.}
        \label{tab:ablation_additional}
        \scriptsize
        \setlength{\tabcolsep}{3pt} %
\setlength{\aboverulesep}{0pt}
\setlength{\belowrulesep}{0pt}
\renewcommand{\arraystretch}{1.4}
\begin{NiceTabular}{l|ccc|cc}
\toprule
Method & MPJPE${\downarrow}$ & MPJVE${\downarrow}$ & Jerk${\downarrow}$ & Hand PE${\downarrow}$ & Vis Hand PE${\downarrow}$ \\ 
\midrule
No Auxiliary & 83.44 & 27.56 & 0.163 & 153.17 & 60.57 \\
\midrule
No Root Ori & 87.68 & 26.95 & 0.120 & 154.81 & 66.17 \\
Head & 82.13 & \textbf{25.43} & 0.119 & 148.71 & 58.17 \\
\midrule
Default & \textbf{81.35} & 25.81 & \textbf{0.117} & \textbf{146.71} & \textbf{56.98} \\
\bottomrule
\end{NiceTabular}

    \end{minipage}
    \hfill
    \begin{minipage}[t]{0.42\textwidth}
        \centering
        \caption{\textbf{Comparison with DSPoser.} Results under DSPoser's $90^\circ$ circular FoV protocol and AvatarPoser split. Since DSPoser does not model body shape, ours uses a fixed mean shape here. DSPoser numbers are from its paper.}
        \label{tab:dsposer}
        \scriptsize
        \setlength{\aboverulesep}{0pt}
\setlength{\belowrulesep}{0pt}
\renewcommand{\arraystretch}{1.462}
\begin{NiceTabular}[t]{l | cc} %
    \toprule
    Method & MPJPE$\downarrow$ & MPJVE$\downarrow$ \\
    \midrule
    DSPoser & 55.1 & 24.19 \\
    Ours & \textbf{52.15} & \textbf{17.58} \\
    \bottomrule
\end{NiceTabular}

    \end{minipage}
\end{table}

\section{Timestep-Adaptive Guidance Optimization}
We provide the objective for the optional timestep-adaptive guidance used during DDIM sampling.
This guidance is used only as a lightweight refinement for visible wrist alignment.
At each sampling step, we minimize $\mathcal{L}_\mathrm{opt}$.
{
\small
\begin{equation}
    \mathcal{L}_\mathrm{opt} = \sum_{t=1}^{T} \sum_j
     \left[ \sqrt{\bar\alpha_n}\, d\big(\tilde{\mathbf{W}}_t^j, \hat{\mathbf{W}}_t^j\big) + s \!\cdot\! v_t^j \!\cdot\! \sqrt{1 - \bar{\alpha}_n}\, d\big(\tilde{\mathbf{W}}_t^j, \mathbf{P}_t^j\big) \right]
\end{equation}
}
Here, $\tilde{\mathbf{W}}_t^j=\mathrm{FK}_j(\tilde{\mathbf{M}}_t^0,\hat\beta)$ and $\hat{\mathbf{W}}_t^j=\mathrm{FK}_j(\hat{\mathbf{M}}_t^0,\hat\beta)$ are the SE(3) poses of hand joint $j\in\{\mathrm{lHand},\mathrm{rHand}\}$ in the refined motion $\tilde{\mathbf{M}}^0$ and the denoiser prediction $\hat{\mathbf{M}}^0$, $\mathbf{P}_t^j$ is the observed wrist pose in world coordinates from $\mathbf{H}$, $d(\mathbf{A},\mathbf{B})=\|\log(\mathbf{A}^{-1}\mathbf{B})^\vee\|^2$ is the squared distance between two rigid transformations, and $s=30.0$ is the guidance scale.
We optimize only the arm joints~\cite{jiang2022avatarposer}.
The first term keeps the motion close to the denoiser prediction, while the second pulls visible wrists toward the observations.
The weights $\sqrt{\bar\alpha_n}$ and $\sqrt{1-\bar\alpha_n}$ make the constraint dominate early steps and the prior dominate late steps, keeping the motion on the manifold.
Because the denoiser already provides a strong hand-aware motion prior, a few LM iterations are sufficient to refine visible wrist alignment from a plausible initialization.
This frame-independent guidance is highly parallelizable and avoids slow sequence-level post-hoc fitting.

\section{Additional Ablation Studies}
We provide additional ablations for training losses and root-orientation representation, complementing the main ablations in Sec.~\ref{sec:ablation}.

\paragraph{Auxiliary Loss}

We validate the auxiliary loss $\mathcal{L}_\mathrm{aux}$ used in Sec.~\ref{sec:augmentation}.
Many diffusion-based motion models~\cite{yi2025egoallo,li2023egoego,patel2025uniegomotion,castillo2023bodiffusion} optimize only the denoising loss on joint rotations, but small rotation errors can accumulate along the kinematic chain and lead to large end-effector errors.
Our auxiliary loss directly regularizes 3D joint positions and foot contacts, making it particularly important for satisfying intermittent hand observations.
Removing $\mathcal{L}_\mathrm{aux}$ degrades both full-body and end-effector accuracy, increasing Vis Hand PE from 56.98~mm to 60.57~mm, and causes a 39\% increase in Jerk.
This confirms that $\mathcal{L}_\mathrm{aux}$ acts as a geometric regularizer that improves hand alignment and motion smoothness.

\paragraph{Root Representation}
We also ablate the root-orientation representation.
The \textbf{No Root Ori} variant mimics EgoAllo~\cite{yi2025egoallo} by directly stitching the predicted body to $\mathbf{T}^\mathrm{head}$, making the root orientation fully determined by the head trajectory.
This removes 3DoF of flexibility and leads to unstable or inaccurate body orientation.
The \textbf{Head} variant restores flexibility by predicting $\mathbf{R}_t^{\text{root}\to\text{head}}$ in head coordinates, but this requires learning motion relative to a volatile head frame.
In contrast, our gravity-aligned canonical representation predicts $\mathbf{R}_t^{\text{root}\to\text{cano}}$, providing a stable reference frame while preserving root-orientation flexibility.
This yields the best balance between stability and expressiveness.

\section{Comparison with DSPoser}
\label{sec:supp_dsposer}

While DSPoser~\cite{chi2024estimatingegobodyposedoubly} is closely related to our work, its official implementation is not publicly available, precluding a direct evaluation under the diverse camera setups investigated in our main text. To ensure a fair and comprehensive comparison, we provide additional results by training and evaluating our model strictly following DSPoser's official protocol and comparing it against their reported metrics. Specifically, we evaluate our framework under their fixed $90^\circ$ circular field-of-view (FoV) setting using the AvatarPoser~\cite{jiang2022avatarposer} train/test split. Furthermore, since DSPoser does not model varying body proportions, we disable our shape prediction module ($\mathcal{E}_\mathrm{shape}$) and enforce a fixed mean body shape during both training and inference to ensure an exact experimental match.

As shown in Tab.~\ref{tab:dsposer}, our device-agnostic framework outperforms the $90^\circ$-trained specialist (DSPoser) even within its native $90^\circ$ domain. This is consistent with our findings in Sec.~\ref{sec:result_fixed}, where our model also outperformed FoV-specific specialist baselines even in their native settings. By exposing our network to a continuous spectrum of simulated visibilities, the model effectively mitigates FoV-specific bias and reduces the geometry overfitting typically exhibited by specialist models trained on fixed configurations.

\section{Additional Qualitative Results for Camera-Setup Generalization}
\begin{figure}[t]\centering
\includegraphics[width=\linewidth, trim={0 0 0 0},clip]{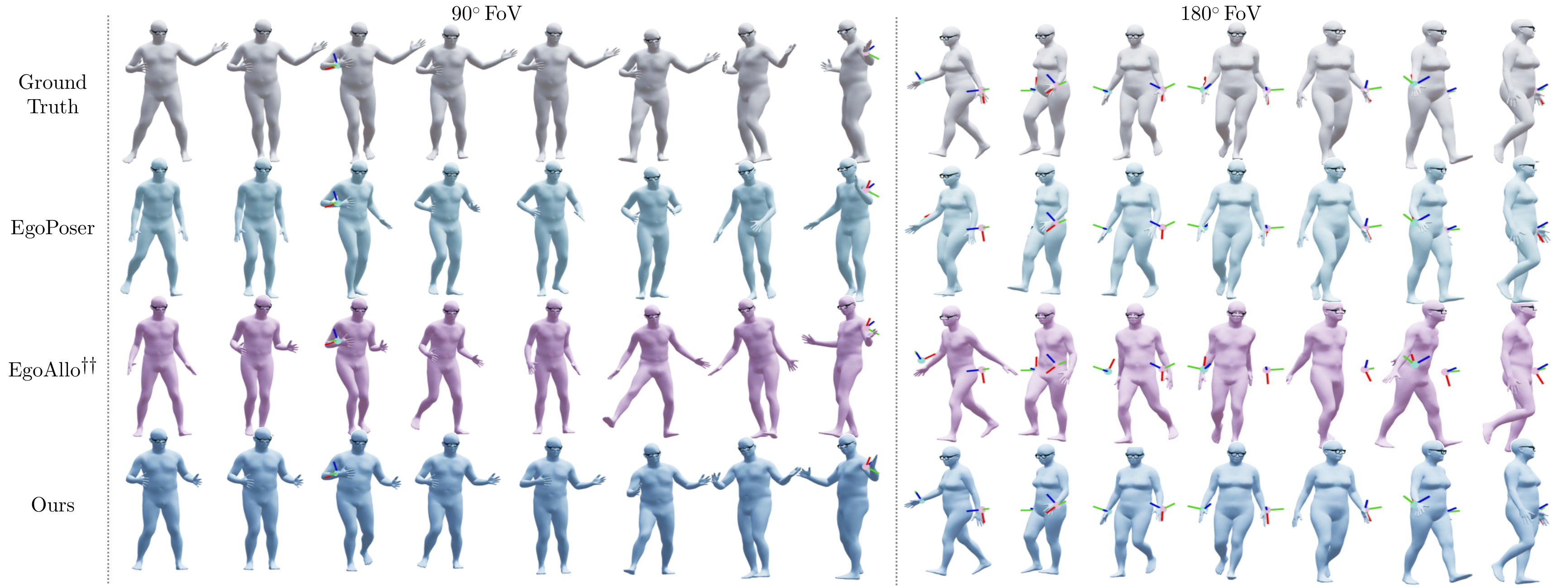}
\caption{\textbf{Qualitative results across varying camera setups.} \textbf{(Left)} Sudden arm movements in EgoPoser (120) and arms pushed out of bounds in EgoAllo$^{\dagger\dagger}$ at $90^\circ$ FoV. \textbf{(Right)} Misaligned tracking of EgoAllo$^{\dagger\dagger}$ at $180^\circ$ FoV.}
\label{fig:qual_fixed}
\end{figure}

Fig.~\ref{fig:qual_fixed} provides qualitative examples complementing the camera-setup generalization results in Sec.~\ref{sec:result_fixed}.
The figure visualizes how fixed-setup specialists fail when the visibility boundary differs from their training regime.
In narrow-FoV settings, EgoAllo$^{\dagger\dagger}$ tends to pull invisible hands toward its learned $120^\circ$ boundary, while in wider-FoV settings it fails to fully exploit hand observations outside this boundary.
EgoPoser (120) can produce low hand-position error in some frames by collapsing invisible hands toward a conservative mean pose, but this leads to discontinuous arm motion when observations appear or disappear.
In contrast, \modelname adapts to the changed camera geometry and maintains plausible hand trajectories and smooth full-body motion.
These examples support the quantitative trends in Table~\ref{table:quant_fixed}, showing that robust generalization requires modeling camera-induced visibility rather than memorizing a fixed observation regime.

\section{Qualitative Analysis of Sequence-Level Context}
As discussed in Sec.~\ref{sec:ablation}, Fig.~\ref{fig:qual_ablation} highlights the role of the visibility context $\bm{\mathcal{G}}$ in resolving out-of-FoV hand ambiguities.
In the initial frames, our full model correctly reconstructs the hands outside the camera's view, whereas the \textbf{No Global} baseline incorrectly places them in front of the body.
As the motion progresses and the wrists later enter the frontal FoV, the full model uses this sequence-level evidence to infer that the initially unobserved hands were outside the camera's observation boundary.
In contrast, \textbf{No Global} lacks this broader temporal context and hallucinates invisible hands within the visible region.
This supports our claim that $\bm{\mathcal{G}}$ captures the visibility context from visibility transitions across the full sequence.

\begin{figure}[t]
    \centering
    \begin{minipage}{0.4755\textwidth}
        \centering
        \input{figures/09_qual_ablation/figure}
        \caption{\textbf{Qualitative Analysis of Ablation Studies.} In a jumping and running sequence, the \textbf{No Global} model incorrectly places invisible hands within the visible zone due to the lack of visibility context. In contrast, our full model correctly infers that the invisible hands are outside the FoV.}
        \label{fig:qual_ablation}
    \end{minipage}
    \hfill
    \begin{minipage}{0.4845\textwidth}
        \centering
        \input{figures/08_representation/figure}
        \caption{\textbf{Representation Visualization.} Illustration of the gravity-aligned canonical frame and the relative geometric transformations used to construct our model's features.}
        \label{fig:representation}
    \end{minipage}
\end{figure}

\section{Linear Probing of the Visibility Context}
\label{sec:supp_probing}
We pair each of the 452 held-out AMASS test motions with 12 cameras sampled independently from our augmentation, giving 5,424 samples, and extract $\bm{\mathcal{G}}$ with the frozen encoder.
The probe target is the camera's visible region over head-relative directions, discretized into 2,592 cells on the unit sphere.
We fit a ridge regression from $\bm{\mathcal{G}}$ to the per-cell visibility and report the area-weighted IoU between the predicted and true regions, computed over the cells within $5^\circ$ of a direction visited by the wrists (19.4\% of the sphere).
We use 5-fold cross-validation with folds grouped by motion, and select the ridge strength and the IoU threshold on a motion-grouped split of the training folds.
The baselines follow the same protocol.
Shuffled labels fits the probe to targets shuffled across samples, mean training camera predicts the visible region of the average training camera, and coarse visibility statistics replaces $\bm{\mathcal{G}}$ with four per-sequence features, namely the visible fraction, its square, the rate of visibility transitions, and the log sequence length.
For stature, we fit the same ridge probe from $\hat\beta$ or $\bm{\mathcal{G}}$ to the wearer's height with leave-one-subject-out cross-validation over the 36 test subjects and report the mean absolute error.

\section{OmniEgoDB Benchmark Details}
\label{sec:supp_omniegodb}
\begin{figure}[t]\centering
\includegraphics[width=\linewidth, trim={0 0 0 0},clip]{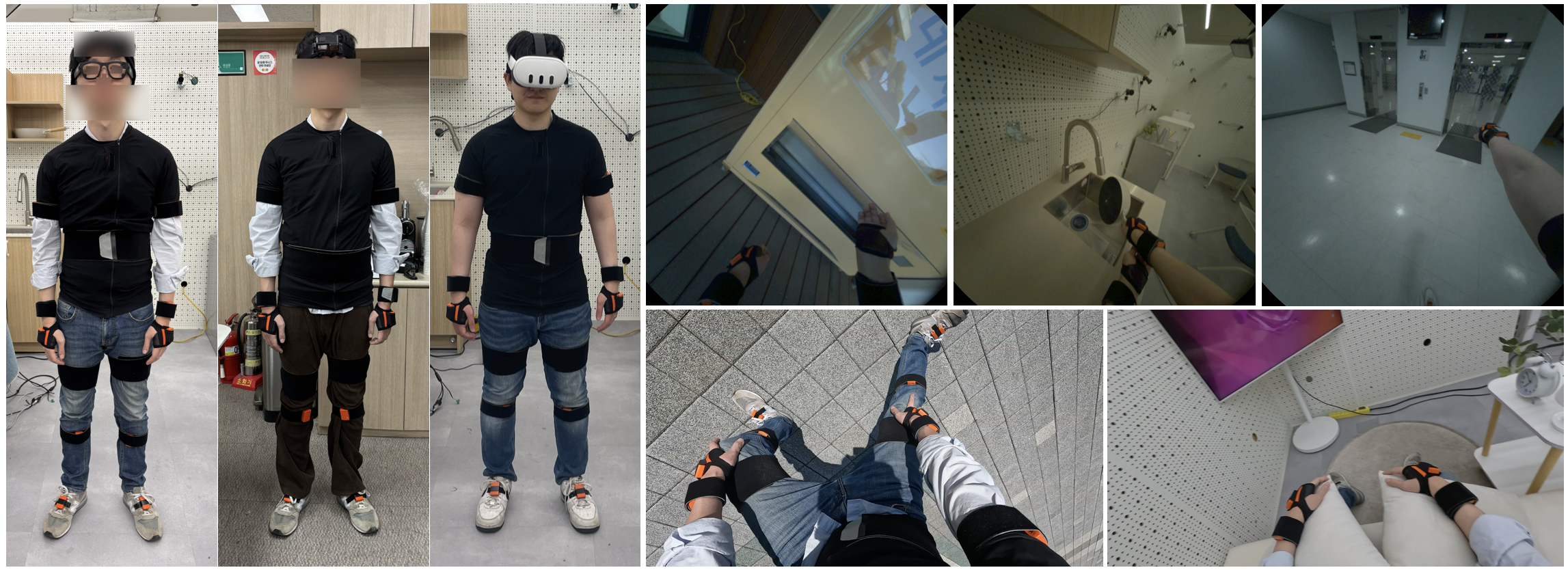}
\caption{\textbf{Overview of OmniEgoDB.} The left side shows participants equipped with three different egocentric devices and Xsens mocap suits.
The right side shows representative egocentric frames from OmniEgoDB, illustrating diverse viewpoints, scenes, and activities.}
\label{fig:omniegodb}
\end{figure}

\begin{figure}[t]\centering
\includegraphics[width=\linewidth, trim={0 0 0 0},clip]{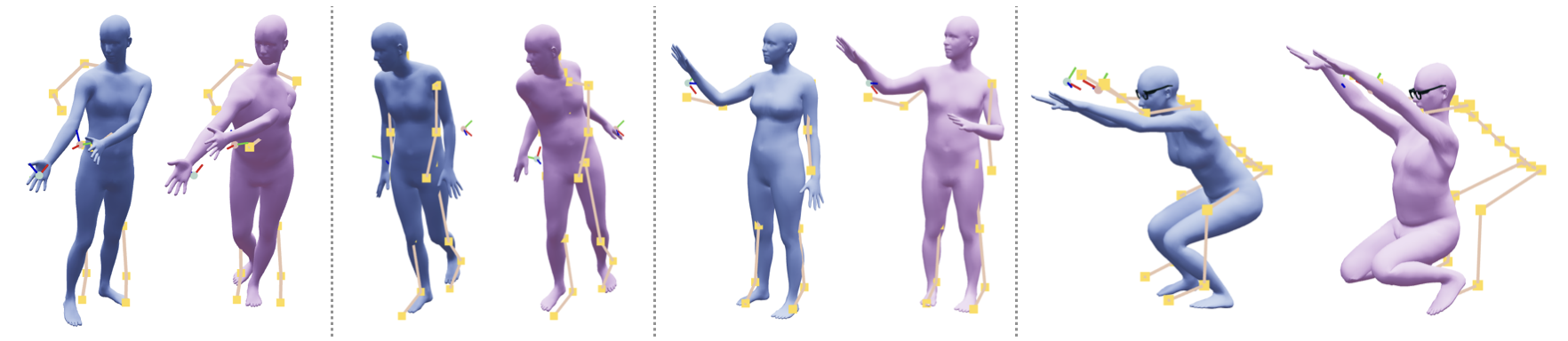}
\captionof{figure}{\textbf{Qualitative Analysis on OmniEgoDB.} From left to right, we show a GoPro depth-direction wrist error, a Quest false-positive wrist observation, temporarily missing Aria wrist observations, and an Aria case with inconsistent body shape estimation.
Blue, purple, and yellow denote \modelname, EgoAllo with optimization, and ground truth, respectively.}
\label{fig:qual_omniegodb}
\end{figure}

In Sec.~\ref{sec:real_world}, we introduce OmniEgoDB as a real-device mocap benchmark for evaluating device-agnostic egocentric motion reconstruction.
Here, we provide additional details on the capture setup, processing and calibration, and qualitative analysis.

\paragraph{Capture Setup}
OmniEgoDB pairs synchronized egocentric videos with Xsens bodysuit ground-truth motion.
We capture natural daily activities using Project Aria~\cite{engel2023projectarianewtool}, GoPro~\cite{gopro}, and Quest~\cite{metaquest}.
These devices represent different egocentric sensing regimes.
Aria provides a smart-glasses setup with built-in perception services, Quest provides an XR headset setup with onboard VIO and real-time hand tracking, and GoPro provides an RGB-only action-camera setup without native head or hand tracking.
The benchmark contains 50 sequences from four subjects totaling 177.2 minutes, with 21 Aria sequences (67.1 min), 20 GoPro sequences (68.1 min), and 9 Quest sequences (42.0 min). The sequences are captured at three indoor locations (an office, a hallway, and the OmniRobotHome platform~\cite{lee2026omnirobothome}) and two outdoor locations (a plaza and a parking lot).
The activities cover locomotion, object manipulation, and dynamic motions, including walking, jumping, cooking, dancing, and exercising.
For GoPro captures, we do not fix the head-mount tilt angle, allowing natural variation in camera mounting.

\paragraph{Processing and Calibration}
OmniEgoDB is processed to provide the input signals required by \modelname, including head trajectories, intermittent hand visibility, and 6-DoF wrist observations.
The processing pipeline differs across devices depending on the available tracking signals.
For Aria, we use Project Aria's Machine Perception Services to obtain the head trajectory and hand tracking signals.
For Quest, we use the headset's VIO-based head tracking and real-time hand tracker.
For GoPro, which does not provide native head or hand tracking, we estimate the camera trajectory with GlueMap~\cite{pan2026gluemap}, recover metric scale from the wearer's camera height, and estimate wrist observations with HaWoR~\cite{zhang2025hawor}.
For Aria (RGB), we undistort the RGB stream to a pinhole camera and estimate wrist observations with the same pipeline, while keeping the MPS head trajectory.

All captures are performed with Xsens suits and gloves to obtain ground-truth full-body motion.
The gloves can partially cover the back of the hands and reduce the reliability of image-based hand detection.
This is especially challenging for GoPro and Aria (RGB), where hand visibility and wrist observations must be estimated from RGB video alone.
We therefore use SAM~3~\cite{carion2026sam}, prompted with ``hand'', to obtain hand bounding boxes for HaWoR.

We temporally synchronize each egocentric recording with the corresponding Xsens motion sequence by estimating the temporal offset that best aligns the magnitudes of the egocentric device and Xsens head angular velocities.
Following the calibration protocol used in Nymeria~\cite{ma2024nymeriamassivecollectionmultimodal}, we estimate a fixed hand-eye transform between each egocentric device trajectory and the Xsens head frame.
Since Xsens motion can accumulate drift over time, we use the egocentric camera trajectory as the reference coordinate system and align the Xsens motion to this camera/world frame.
This yields a common coordinate system for evaluating reconstructed SMPL-H motion against Xsens ground truth.

\paragraph{Analysis of Benchmark Results}
Table~\ref{table:quant_benchmark} shows consistent gains across Aria, GoPro, and Quest, with different error patterns across baselines.
EgoAllo is sensitive to inaccurate or jittery wrist constraints, especially on GoPro, where both camera motion and wrist observations must be estimated from RGB video.
Its optimization can propagate these errors to the arms or whole body, producing implausible configurations such as body interpenetration or twisted arms.
In contrast, \modelname relies on a hand-aware diffusion prior and sequence-level visibility context, maintaining stable full-body motion while optional guidance mainly improves Hand PE.

Figure~\ref{fig:qual_omniegodb} further compares \modelname with EgoAllo under noisy or unreliable wrist observations.
In the GoPro example, a large depth-direction wrist error causes EgoAllo to overfit the erroneous constraint and produce body interpenetration, whereas \modelname yields a plausible reaching motion.
In the Quest example, a transient false-positive wrist observation is corrected in the following frames, and \modelname uses this temporal context to suppress the outlier, while EgoAllo remains over-constrained by the erroneous signal.
The two Aria examples show complementary failures of EgoAllo.
When wrist observations are temporarily missing, EgoAllo hallucinates unstable arm motion.
In another case, inconsistent body shape estimation distorts the whole-body configuration and leads to an incorrect squat motion.
These examples highlight that \modelname is more robust to transient tracking errors by leveraging sequence-level visibility context and a consistent body shape.

\section{Evaluation on Aria Digital Twin}
\label{sec:supp_adt}
We evaluate all methods zero-shot on Aria Digital Twin (ADT)~\cite{pan2023ariadigitaltwin}, which pairs Aria recordings with OptiTrack body motion.
We use 34 of the 35 single-person recordings, excluding one whose MPS hand tracking is unavailable, and score the 15 joints shared by SMPL-H and the OptiTrack skeleton at 30 FPS.
As for OmniEgoDB, ADT (Stereo) uses the MPS head trajectory and hand tracking, and ADT (RGB) uses HaWoR~\cite{zhang2025hawor} wrists from the RGB stream with the same head trajectory.
Because the OptiTrack head joint lies at a different point than the SMPL-H head, we shift the head trajectory given to our model and EgoPoser by a fixed offset estimated once from three recordings.
Table~\ref{table:adt} shows that \modelname achieves the lowest error on all four metrics in both settings, and EgoAllo with optimization remains 18--24~mm behind in MPJPE.
The official EgoPoser comes close in MPJPE and is more accurate in recordings where the wearer is mostly seated, such as desk work, which we attribute to its regression toward the conditional mean pose fitting static seated postures.
Its motion is, however, far less smooth, with roughly twice the MPJVE and about seven times the Jerk of ours in ADT (Stereo), whereas \modelname produces smooth and natural motion.

\begin{table}[t]
\centering
\caption{
\textbf{Quantitative results on Aria Digital Twin.}
$^\#$ denotes methods evaluated with their official checkpoints.
-- denotes entries not applicable to UniEgoMotion, which needs RGB frames and predicts at 10 FPS.
}
\label{table:adt}
\scriptsize
\setlength{\tabcolsep}{2.2pt}
\setlength{\aboverulesep}{0pt}
\setlength{\belowrulesep}{0pt}
\renewcommand{\arraystretch}{1.4}
\begin{NiceTabular}{X[2,l]|X[c]X[c]X[c]X[c]|X[c]X[c]X[c]X[c]}[width=\textwidth]
\toprule
\multirow{3}{*}{Method}
& \multicolumn{4}{c|}{ADT (Stereo)}
& \multicolumn{4}{c}{ADT (RGB)} \\[-3.5pt]
& \multicolumn{4}{c|}{\tiny 34 seqs, 52.5 min}
& \multicolumn{4}{c}{\tiny 34 seqs, 52.5 min} \\
\cmidrule{2-9}
& MPJPE$\downarrow$ & MPJVE$\downarrow$ & Jerk$\downarrow$ & Hand PE$\downarrow$
& MPJPE$\downarrow$ & MPJVE$\downarrow$ & Jerk$\downarrow$ & Hand PE$\downarrow$ \\
\midrule
EgoAllo$^\#$ (w/o opt)
& 159.06 & 18.36 & 0.117 & 293.74
& 159.06 & 18.36 & 0.117 & 293.74 \\
EgoAllo$^\#$ (w/ opt)
& 145.78 & 19.61 & 0.075 & 212.97
& 150.04 & 26.08 & 0.256 & 204.78 \\
EgoPoser$^\#$
& 127.50 & 32.22 & 0.429 & 236.36
& 126.86 & 46.84 & 0.950 & 206.69 \\
EgoPoser (120)
& 131.06 & 40.20 & 0.857 & 243.56
& 130.56 & 61.49 & 1.511 & 213.93 \\
UniEgoMotion$^\#$
& -- & -- & -- & --
& 141.99 & -- & -- & 232.31 \\
\midrule
Ours
& 127.14 & \textbf{17.49} & \textbf{0.062} & 193.10
& 125.86 & \textbf{17.66} & \textbf{0.067} & 199.55 \\
Ours (w/ opt)
& \textbf{125.40} & 17.91 & 0.074 & \textbf{189.80}
& \textbf{125.63} & 19.76 & 0.144 & \textbf{197.32} \\
\bottomrule
\end{NiceTabular}
\end{table}

\section{Representation Visualization}
We provide visual illustrations of the coordinate representations detailed in Sec.~\ref{sec:preprocessing_representation} of the main text.
Following EgoAllo~\cite{yi2025egoallo}, the canonical frame serves as a gravity-aligned local reference. It is defined by projecting the head coordinate onto the ground plane, aligning its \(z\)-axis with gravity and its \(y\)-axis with the head's forward direction. 
Based on this frame, Fig.~\ref{fig:representation} visualizes the spatio-temporally invariant relative transformations used to construct our model's features. These include the relative head trajectory ($\Delta\mathbf{T}_t^\text{head}$), wrist positions relative to the head ($\mathbf{T}_t^{\text{hand}\to\text{head}}$), canonicalized head orientation ($\mathbf{R}_t^{\text{head}\to\text{cano}}$), head height ($h_t$), relative canonical rotation ($\Delta\mathbf{R}_t^\text{cano}$), and relative root orientation ($\mathbf{R}_t^{\text{root}\to\text{cano}}$).

\section{Calibration and World Alignment}
\label{sec:supp_preprocessing}
We present the detailed calibration and alignment procedures used to map raw tracking data into the gravity-aligned world coordinate system, expanding upon Sec.~\ref{sec:preprocessing_representation} of the main text. Importantly, these steps are lightweight one-time preprocessing procedures and do not require any device-specific retraining.

\paragraph{Camera-to-Head Calibration} 
To map the camera trajectory to the head joint, we must determine the rigid transformation $\mathbf{T}^{\text{cam}\to\text{head}}$. For Aria~\cite{metaaria} and Quest~\cite{metaquest}, which track the center eye frame, we use a precomputed transformation based on a mean body shape. For monocular cameras, estimating the absolute metric translation between the camera and the head joint is ill-posed. Therefore, we decouple the calibration of translation and orientation. We approximate the translation offset based on the device mounting configuration (\eg, using an external reference image). For the orientation, we use a brief pre-calibration phase where the user stands straight and looks forward. Since this posture approximately aligns with our predefined canonical head frame, we can estimate the relative rotation from the camera to the head joint from this initialization.

\paragraph{Ground Plane Fitting and Scale Correction}
To establish the world coordinate system, we determine the ground plane by extracting the global 3D point cloud—generated either by the device's native SLAM system (\eg, Aria) or the DROID-SLAM~\cite{teed2021droid} backend within a modified HaWoR pipeline~\cite{zhang2025hawor}. We apply RANSAC to this point cloud to fit the ground plane, using its normal to define the gravity-aligned $z$-axis. 

For monocular in-the-wild captures, we further refine the global scale for qualitative visualization. Since the initial metric scale from HaWoR is often inaccurate, we estimate a single global scale correction factor during the static standing phase of the pre-calibration by comparing the estimated camera height against an approximate physical camera height based on the user's height and device placement. This correction is used only as a preprocessing step to resolve the metric ambiguity of monocular SLAM for qualitative in-the-wild examples.

\section{Additional Qualitative Results in the Wild}
\begin{figure}[t]\centering
\includegraphics[width=\linewidth, trim={0 0 0 0},clip]{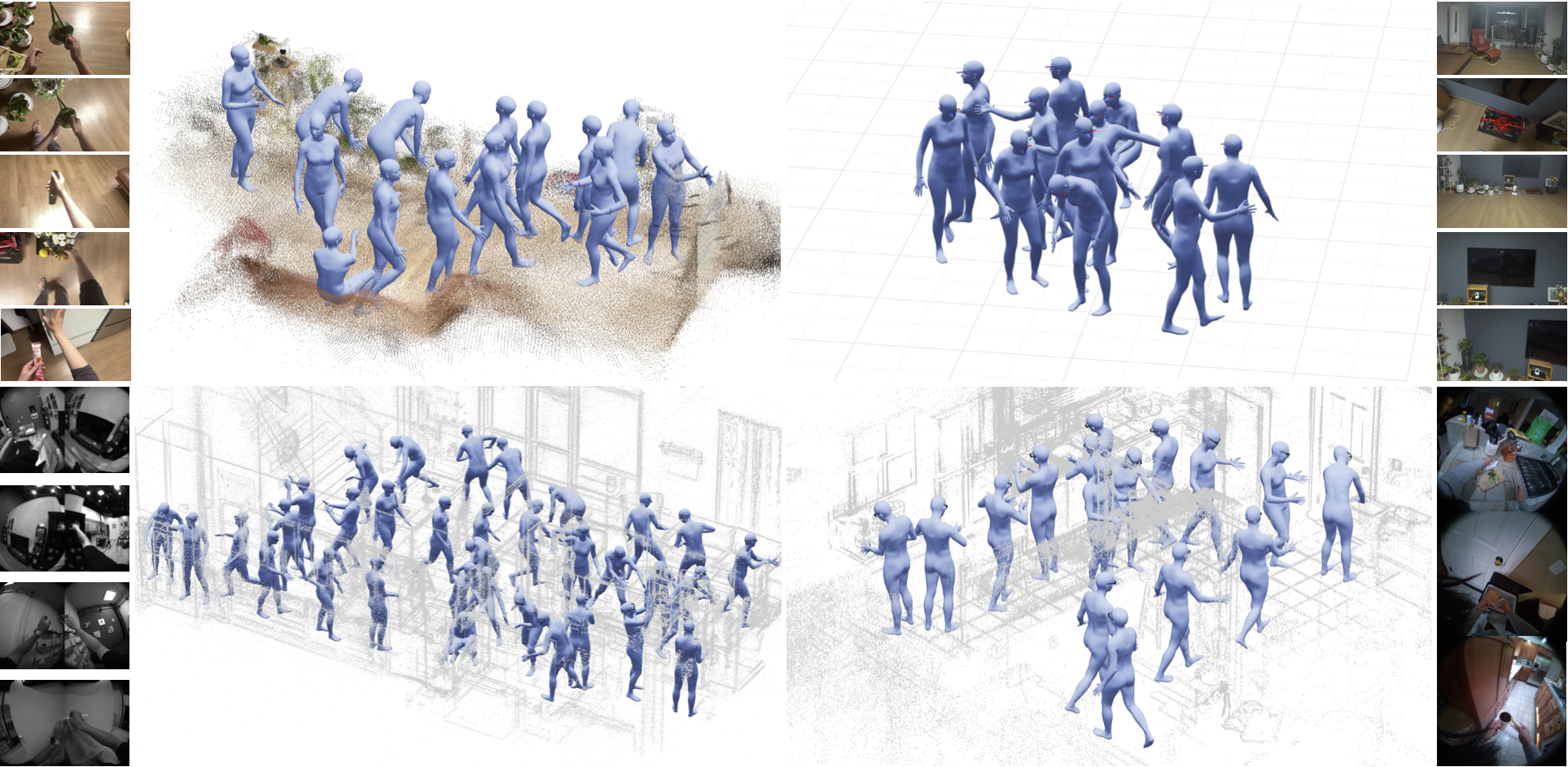}
\captionof{figure}{\textbf{Additional qualitative results on in-the-wild data.} Reconstructions from (Upper Left) iPhone 15, (Upper Right) Quest, (Lower Left) Aria Stereo (SLAM), and (Lower Right) Aria Monocular (RGB).}
\label{fig:in_the_wild_additional}
\end{figure}

In Fig.~\ref{fig:in_the_wild_additional}, we provide additional in-the-wild reconstruction results captured across diverse setups, including an iPhone 15 (26mm lens), Meta Quest 3~\cite{metaquest}, and Project Aria~\cite{metaaria}. Notably, for the Aria RGB sequences, we determine hand visibility by projecting the device's 3D hand tracking outputs onto the 2D image plane. For the Quest captures, the visualized front RGB image is shown only for reference; hand tracking is estimated by the headset's onboard camera system rather than being restricted to the displayed RGB view. Complementing the main text, these diverse examples further highlight \modelname's robust and device-agnostic generalization capabilities across unconstrained camera setups. We encourage readers to visit our project page for animated visualizations.

\section{Failure Cases}
\label{sec:failure_cases}
\begin{figure}[t]\centering
\includegraphics[width=\linewidth, trim={0 0 0 0},clip]{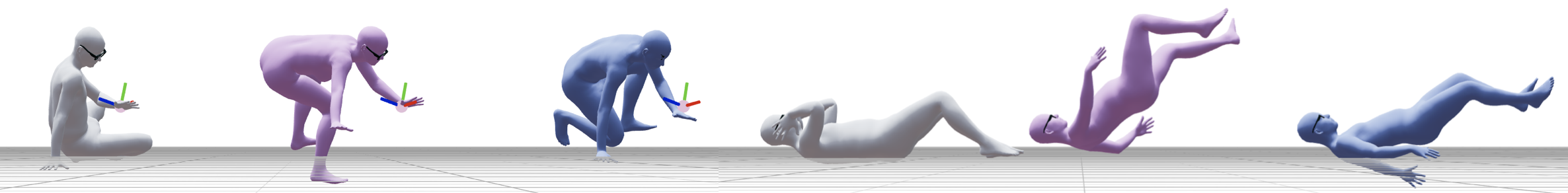}
\caption{\textbf{Representative failure cases.} Gray corresponds to the ground truth, blue to Ours (w/ opt), and purple to EgoAllo (w/ opt).}
\label{fig:failure}
\end{figure}

We further investigate representative failure cases in Fig.~\ref{fig:failure}, which are shared by head-centric egocentric reconstruction methods, including both EgoAllo~\cite{yi2025egoallo} and our framework. First, seated poses remain inherently ambiguous in an egocentric view, as similar head and hand trajectories may correspond to squatting, sitting on the floor, or sitting on a chair. Second, near-lying poses may introduce minor foot-floating artifacts, as the head-centric canonicalization becomes less stable in such extreme configurations. Additionally, our reconstruction quality may also degrade when the upstream SLAM or hand tracking signals become unreliable.

\end{document}